\documentclass{article}
\usepackage[utf8]{inputenc}
\usepackage[hidelinks]{hyperref}
\usepackage{amsmath,amsfonts,amsthm,amssymb}
\usepackage{mhchem}
\usepackage{gensymb}
\usepackage{nicefrac}
\usepackage{geometry}
\usepackage{graphicx}
\usepackage{xcolor}
\usepackage{tikz}
\usetikzlibrary{positioning, shapes.geometric, shapes.symbols, shapes.arrows, calc, fit, decorations.pathreplacing}
\usetikzlibrary{positioning, shapes.geometric, shapes.symbols, shapes.arrows, calc, fit}
\usepackage{authblk}
\definecolor{mygreen}{RGB}{50, 180, 50}

\usepackage{booktabs}
\usepackage{multirow}
\usepackage{caption}
\usepackage{subcaption}

\usepackage[displaymath, mathlines]{lineno}

\title{\textbf{Machine learning in wastewater treatment: insights from modelling a pilot denitrification reactor}}
\author[1]{Eivind Bøhn}
\author[1]{Sølve Eidnes\thanks{Corresponding author}}
\author[2]{Kjell Rune Jonassen}

\affil[1]{\small{Department of Mathematics and Cybernetics, SINTEF Digital, 0373 Oslo, Norway}\par
\texttt{\{eivind.bohn, solve.eidnes\}@sintef.no}}
\affil[2]{\small{Veas AS, Eternitveien 125, 3470 Slemmestad, Norway}\par \texttt{krj@veas.nu}}

\date{June 28, 2025}

\begin{document}
\maketitle

%\linenumbers

\begin{abstract}
Wastewater treatment plants (WWTPs) are promising candidates for machine learning (ML) due to their societal relevance and data availability. However, differences in plant design, operation, and influent characteristics hinder the generalisation of ML models across sites. More research is needed to understand how to tackle these challenges. This study analyses a pilot denitrification reactor at the Veas facility in Norway. The controlled conditions and high-resolution sensing of the pilot provide a valuable complement to modelling based on long-term process data. Rather than maximizing predictive accuracy, we address foundational requirements for effective data-driven modelling of WWTPs:\ identifying critical input parameters, assessing data sufficiency and structure, and examining model behaviour. Our results show nonlinear models fit training and validation data best, suggesting learnable nonlinearities, but linear models generalise better to an unseen test period. A key challenge is the temperature variable, whose seasonal shift between training and test periods significantly impairs performance. This highlights the need for multi-year datasets to robustly capture climate-induced variability, especially in northern regions. By emphasising model interpretability, data requirements, and generalisation limits, we offer practical insights for future ML deployment in WWTPs. We publicly share the data and code used to produce the results.
\end{abstract}

\vspace{6pt}

\section{Introduction}\label{sec:introduction}
The Oslofjord, a 120 km long inlet in southeastern Norway, serves as an oceanic connection for approximately two million people. Unfortunately, the populations of phytoplankton, fish and marine bird populations in and around the fjord are decreasing. This decline is hypothesised to be caused by a high input of nutrients and carbon, leading to poor oxygen availability with consequences for the fjord’s ecological status. One significant driver of oxygen deficiency is the increasing amount of reactive nitrogen from agricultural fertilizer runoff \cite{staalstrom2023undersokelse} and wastewater treatment plants (WWTP). In 2022 WWTPs alone contributed to 24\% of the input of reactive nitrogen to the outer Oslofjord \cite{engesmo2023overvaaking}. Efficient nitrogen removal at these plants is therefore assumed crucial for improving the ecological status of the fjord.

The Oslofjord receives wastewater, directly and indirectly via rivers, from 118 municipalities, with a combined population of 2.8 million people -- more than half the total population of Norway and a million more than in the early 1990s. A shallow threshold restricts water circulation in the fjord's inner parts, and by the late 1990s, population growth necessitated the implementation of biological nitrogen removal processes in wastewater treatment plants discharging into these areas. However, similar treatment initiatives were not implemented at WWTPs discharging into the outer parts of the fjord \cite{engesmo2023overvaaking}.

Today, five WWTPs releasing water directly or indirectly to the Oslofjord have biological nitrogen removal processes implemented. With increased awareness of environmental impacts, and the implementation of stricter regulations \cite{european_parliament2024directive}, %\footnote{See the European Parliament legislative resolution of 10 April 2024 on the proposal for a directive of the European Parliament and of the Council concerning urban wastewater treatment (recast) (COM(2022)0541 – C9-0363/2022 – 2022/0345(COD)).}, 
more treatment plants will adopt nitrogen removal processes in the near future. This motivates finding ways to make the removal of nitrogen as efficient as possible \cite{staalstrom2022utredning}.

The Veas WWTP, located at Slemmestad in the Inner Oslofjord, is the largest WWTP in Norway. The plant serves approximately 640,000 residents and industry in the greater Oslo region, treating a nutrient load equivalent to 800,000 people.  Authorities mandate that Veas, as an annual average, remove 70\% of the nitrogen, 90\% of the phosphorous and 75\% of the organic carbon from the wastewater before releasing it to the fjord.

\begin{figure}[ht!]
    \centering
    \resizebox{\textwidth}{!}{\begin{tikzpicture}[
    node distance=1cm and 0.5cm,
    boxtank/.style={rectangle, draw, line width=1pt, minimum size=0.75cm, align=center},
    nottank/.style={diamond, draw, line width=1pt, minimum size=0.75cm},
    outbox/.style={signal, draw, line width=1pt, minimum size=0.75cm},
    product/.style={signal, signal to=east, shape border rotate=90, draw, line width=1pt, minimum size=0.75cm, align=center},
    bowl/.style={semicircle, shape border rotate=180, draw, line width=1pt, minimum size=0.75cm, align=center},
    spacer/.style={minimum size=0cm, inner sep=0cm, outer sep=0cm},
    bigbox/.style={draw, line width=1pt, rectangle, inner sep=0.25cm},
    >=latex % Specifies the arrow tip style
]

% Define the custom shape as a 'pic'
\tikzset{
  pics/insignal/.style={
    code={
      \coordinate (A) at (0,0);
      \coordinate (B) at (0,-0.4);
      \coordinate (C) at (-1.5,-0.4);
      \coordinate (D) at (-1.1,0);
      \coordinate (E) at (-1.5,0.4);
      \coordinate (F) at (0,0.4);
      % Draw the shape
      \draw[line width=1pt] (A) -- (B) -- (C) -- (D) -- (E) -- (F) -- cycle;
      
      %% Optionally, add text label inside the shape
      \node at ($(D)!0.5!(A)$) {Inlet};
    }
  }
}

% Adding curly brackets above the diagram
\draw[decorate,decoration={brace,amplitude=10pt,raise=2pt}] (-11.7, 2.0) -- (-7.3, 2.0) node[midway,yshift=18pt]{Mechanical treatment};

\draw[decorate,decoration={brace,amplitude=10pt,raise=2pt}] (-9.8, 3.1) -- (-3.6, 3.1) node[midway,yshift=18pt]{Chemical treatment};

\draw[decorate,decoration={brace,amplitude=10pt,raise=2pt}] (-3.1, 3.1) -- (6.0, 3.1) node[midway,yshift=18pt]{Biological treatment};

%% Boxes:

\pic at (-12.25,-1.2) {insignal};

%% Swamp
\node [boxtank] (swamp) at (-11,-1.2) {Screens};

% Waste from swamp
\node [product] (waste) at ($(swamp)-(0,1.6cm)$) {Waste};

%% Sand and grit removal
\node [boxtank] (sand) at (-9,-1.2) {Aerated\\ grit\\ chambers};

% Waste from swamp
\node [product] (sandproduct) at ($(sand)-(0,1.6cm)$) {Sand};

% Chemicals added to sand and grit removal
\node [bowl] (chemicals) at ($(sand)+(0,2.3cm)$) {$\textrm{AlCl}_3$ and \\ $\textrm{FeCl}_3(\textrm{H}_2\textrm{O})_x$};

% Sedimentation
\node [boxtank] (sedimentation) at (-5,0) {Sedimentation};

% Polymers added to sedimentation
\node [bowl] (polymer) at ($(sedimentation)+(0,1.4cm)$) {Polymer};

% Nitrate product from sedimentation
\node [product] (nitrate) at ($(sedimentation)-(0,1.4cm)$) {Sludge};

% Nit filterstanks
\node [boxtank, label=above:Nitrification] (nit-1) at (-2,1.5) {Reactor 1};
\foreach \m in {2,...,4}
  \node [boxtank] (nit-\m) at (-2,2.5-\m) {Reactor \m};
  
% Den-tank
\node [boxtank] (dentank-1) at (0,0) {Buffer\\ volume};

% Pumps
\node [nottank] (pumps-1) at (3,0) {Pumps};

% Methanol added before the pumps
\node [bowl] (methanol) at ($(dentank-1)!0.5!(pumps-1)+(0,1.4cm)$) {Methanol};

% Denit filters
\node [boxtank, label=above:Denitrification] (filters-1) at (5,1.5) {Reactor 1};
\foreach \m in {2,...,4}
  \node [boxtank] (filters-\m) at (5,2.5-\m) {Reactor \m};

% Pilotpump
\node [nottank] (pilotpump) at (4.0,-4.4) {Pump};

% Methanol added before the pumps
\node [bowl] (pilotmethanol) at (1.7,-4.4) {Methanol};

% Pilot filtertanks
\node [boxtank] (pilotfilter) at (6.0,-4.4) {Denitri-\\ fication\\ reactor};

% Outlet
\node [outbox] (output-1) at (7,0) {Outlet};

% Spacer Node above Nitrification and Denitrification labels
\node[spacer] (spacer1) at ($(nit-1.north)!0.5!(filters-1.north) + (0,0.35cm)$) {};
% Spacer Node below Nitrification and Denitrification boxes
\node[spacer] (spacer2) at ($(nit-4.south)!0.5!(filters-4.south) + (0,-0.55cm)$) {};

%% Big Box around the diagram representing Process Hall 1
\node[bigbox, fit=(sedimentation) (nitrate) (output-1) (methanol) (spacer1) (spacer2), label=above:Process line 1] (processhall1) {};
%\node[bigbox, fit=(sedimentation) (nitrate) (output-1) (methanol) (spacer1) (filters-4), label=above:Process line 1] (processhall1) {};

% Boxes for Process Halls 2 to 8 below Process Hall 1
\node[boxtank, below=of processhall1, minimum width=2cm, xshift=-6.19cm, yshift=0.7cm] (processhall2) {Process line 2};
\node[align=center, yshift=0.3cm] at ($(processhall2.south)+(0,-0.5cm)$) (dots) {$\vdots$};
\node[boxtank, below=of dots, minimum width=2cm, yshift=1.0cm] (processhall8) {Process line 8};

%% Big Box around the diagram representing the pilot
\node[bigbox, fit=(pilotmethanol) (pilotpump) (pilotfilter), label=above:Pilot, draw=mygreen, line width=2pt] (pilothall) {};

%% Arrows:

% Arrows from inlet to swamp
\draw[->, line width=1pt] (-12.25,-1.2) -- (swamp);

% Arrows from swamp to sand
\draw[->, line width=1pt] (swamp.east) -- (sand.west);

% Arrow from swamp to waste
\draw [line width=1pt, ->] (swamp) -- (waste);

% Draw an arrow from chemicals to Sand
\draw [line width=1pt, ->] (chemicals) -- (sand);

% Arrow from sand to sand product
\draw [line width=1pt, ->] (sand) -- (sandproduct);

% Arrows from sand to Process Halls
\draw[->, line width=1pt] (sand.east) -- (sedimentation.west);
\draw[->, line width=1pt] (sand.east) -- (processhall2.west);
%\draw[->, line width=1pt] (sand.east) -- (dots.west);
\draw[->, line width=1pt] (sand.east) -- (processhall8.west);

% Draw an arrow from Polymer to Sedimentation
\draw [line width=1pt, ->] (polymer) -- (sedimentation);

% Draw an arrow from Sedimentation to Nitrate product
\draw [line width=1pt, ->] (sedimentation) -- (nitrate);

% Connect sedimentation to nit filters
\foreach \j in {1,...,4}
    \draw [line width=1pt, ->] (sedimentation) -- (nit-\j.west);
    
% Connect nit to den-tank with arrows
\foreach \i in {1,...,4}
  \draw [line width=1pt, ->] (nit-\i.east) -- (dentank-1);
    
% Connect Den-tank to pumps with arrow
\draw [line width=1pt, ->] (dentank-1) -- (pumps-1);

% Methanol added before the pumps
\draw [line width=1pt, ->] (methanol) -- ($(dentank-1)!0.5!(pumps-1)$);
    
% Connect pumps to filters with arrows
\foreach \j in {1,...,4}
    \draw [line width=1pt, ->] (pumps-1) -- (filters-\j.west);

% Connect filters to outlet with arrows
\foreach \i in {1,...,4}
  \draw [line width=1pt, ->] (filters-\i.east) -- (output-1);

% Draw an arrow from Methanol to the line between Storage tank and Pumps
\draw [line width=1pt, ->] (methanol) -- ($(dentank-1)!0.5!(pumps-1)$);

% Arrows from dentank to pump in pilot
\draw[->, line width=1pt] (dentank-1) -- (pilotpump);

% Methanol added before pumps in pilot
\draw [line width=1pt, ->] (pilotmethanol) -- ($(dentank-1)!0.8!(pilotpump)$);

% Arrows from pump to filter in pilot
\draw[->, line width=1pt] (pilotpump) -- (pilotfilter);

% Arrow from pilot filter back to inlet
\draw[-, line width=1pt] (pilotfilter.east) -- ($(pilotfilter.east) + (0.5,0.0)$);
\draw[-, line width=1pt] ($(pilotfilter.east) + (0.5,0.0)$) -- ($(pilotfilter.east) + (0.5,-1.4)$);
\draw[-, line width=1pt] ($(pilotfilter.east) + (0.5,-1.4)$) -- ($(pilotfilter.east) + (-19.74,-1.4)$);
\draw[->, line width=1pt] ($(pilotfilter.east) + (-19.74,-1.4)$) -- (-13.0,-1.6);

% Arrow from nitrification back to inlet
\draw[dashed, line width=1pt] (nit-4.south) + (-0.2,0.0) -- ($(nit-4.south) + (-0.2,-3.6)$);
\draw[dashed, line width=1pt] ($(nit-4.south) + (-0.2,-3.6)$) -- ($(pilotfilter.east) + (-19.44,-1.1)$);
\draw[dashed,->, line width=1pt] ($(pilotfilter.east) + (-19.44,-1.1)$) -- (-12.7,-1.6);

% Line from denitrification to pipe from nitrification
\draw[dashed, line width=1pt] (filters-4.south) + (-0.2,0.0) -- ($(filters-4.south) + (-0.2,-0.6)$);
\draw[dashed, line width=1pt] ($(filters-4.south) + (-0.2,-0.6)$) -- ($(nit-4.south) + (-0.2,-0.6)$);

% Line from outlet to nitrification and denitrificaiton
\draw[dashed, line width=1pt] (output-1.south) -- ($(output-1.south) + (0.0,-1.9)$);
\draw[dashed, line width=1pt] ($(output-1.south) + (0.0,-1.9)$) -- ($(output-1.south) + (-8.8,-1.9)$);
\draw[dashed,->, line width=1pt] ($(output-1.south) + (-8.8,-1.9)$) -- ($(nit-4.south) + (0.2,0.0)$);
\draw[dashed,->, line width=1pt] ($(filters-4.south) + (0.2,-0.4)$) -- ($(filters-4.south) + (0.2,0.0)$);

\end{tikzpicture}}
\caption{A schematic overview of the full wastewater treatment process at Veas, which can be separated into mechanical- chemical-, and biological treatment steps. Dashed lines indicate backwashing lines. The green box highlights the denitrification pilot we are modelling in this paper. Details on this process are given in Appendix \ref{sec:veas_detailed}.}
\label{fig:veas}
\end{figure}
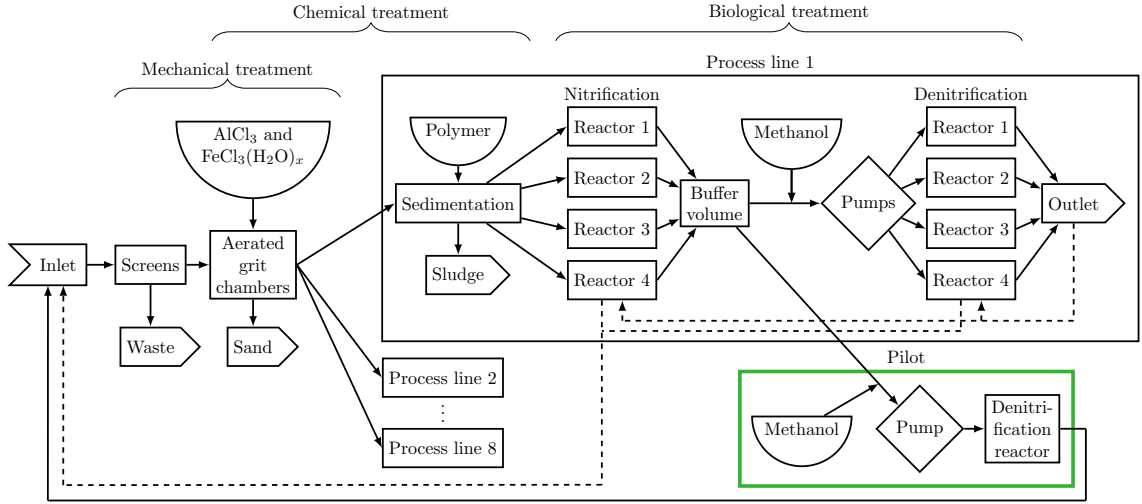

Figure \ref{fig:veas} presents a simplified flow chart of the process at Veas. Nitrogen bound to particulates, along with carbon and phosphorus, is removed through precipitation and sedimentation, while soluble nitrogen and carbon are metabolized via two biogeochemical processes: nitrification and denitrification. Both processes are operated as stationary biofilm systems, using expanded clay as the carrier material. Nitrification is an autotrophic, aerobic process that oxidizes ammonium to nitrate in the presence of oxygen and inorganic carbon. Denitrification is a heterotrophic, anoxic process reducing nitrate to dinitrogen gas, achieved at Veas through the controlled addition of methanol as an external electron donor.

Veas has installed two pilot reactors to experiment with various process optimizations for denitrification. One pilot functions as a reference and is operated similarly to the main process. Data from this pilot has been gathered over a period of 5.5 months and forms the basis of the machine learning (ML) analysis in this paper. While this dataset is from a very limited period, the pilot study provides a controlled environment with consistent sensor data and no feedback loop, in contrast to the process in the full-scale plant. Eventually, the goal is to have a data-driven simulation, or a digital twin, of the actual physical treatment process. An accurate model can then be used to test different strategies and optimise the denitrification process, specifically regarding the amount of added chemicals. Details on the treatment process at Veas and the pilot reactors in particular are given in Appendix \ref{sec:veas_detailed}.

As artificial intelligence (AI), and ML in particular, has been attracting increasingly more interest from both the research community and the general public in recent years, wastewater treatment has stood out as an appealing application area for this technology. Zhang et al.\ \cite{zhang2023artificial} found that the number of yearly research papers on AI for wastewater treatment grew from less than 100 in 2015 to over 600 in 2022. All modern wastewater treatment plants collect data, albeit of varying quality and quantity, for monitoring and operational support. Furthermore, since wastewater treatment generally is a public service, the operations and data of the plant are usually more available for researchers than that of other industries. On the other hand, wastewater treatment is a complicated and challenging process, as exemplified by the description of the Veas process in Appendix \ref{sec:veas_detailed}.

\begin{figure}[ht!]
    \centering
    \resizebox{\textwidth}{!}{
\input{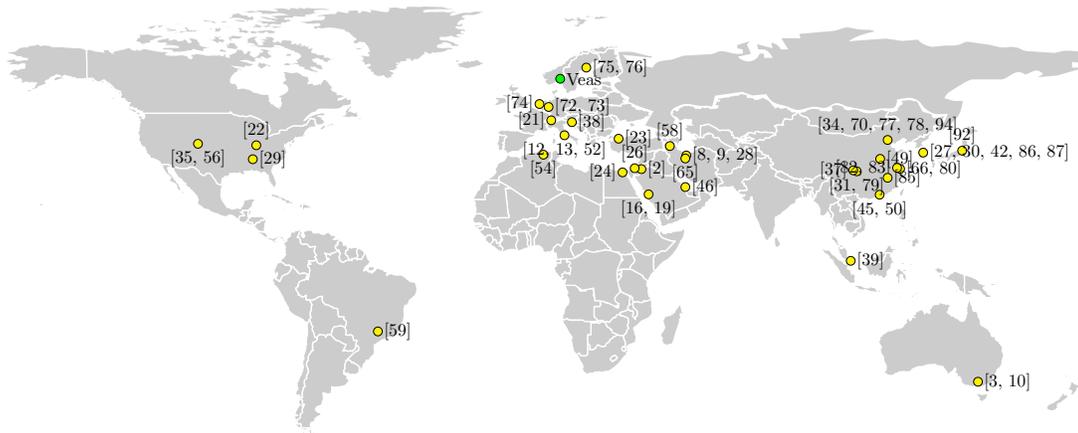}
\begin{tikzpicture}
\WORLD[every state={draw=white, thick, fill=black!20}]

% Oslo
\coordinate (Oslo) at ($(Norway.center)+(-.43,-.57)$);
\draw[fill=green] (Oslo) circle (2.5pt); 
\node[anchor=west] at (Oslo) {\fontsize{10pt}{9.6pt}\selectfont Veas};

% Teheran, Iran
\coordinate (Iran) at ($(Iran.center)+(-.13,.19)$);
\draw[fill=yellow] (Iran) circle (2.5pt); 
\node[anchor=west] at (Iran) {\fontsize{10pt}{9.6pt}\selectfont \cite{bagheri2015modeling, bagherzadeh2021comparative, gholizadeh2024machine}};

% Tabriz, Iran
\coordinate (Tabriz) at ($(Iran.center)+(-.33,.19)$);
\draw[fill=yellow] (Tabriz) circle (2.5pt); 
\node[anchor=south] at (Tabriz) {\fontsize{10pt}{9.6pt}\selectfont \cite{nourani2021artificial}};

% Qom, Iran
\coordinate (Qom) at ($(Iran.center)+(-.02,-.06)$);
\draw[fill=yellow] (Qom) circle (2.5pt); 
\node[anchor=north] at (Qom) {\fontsize{10pt}{9.6pt}\selectfont \cite{sharafati2020potential}};

% Singapore
\coordinate (Singapore) at ($(Singapore.center)$);
\draw[fill=yellow] (Singapore) circle (2.5pt); 
\node[anchor=west] at (Singapore) {\fontsize{10pt}{9.6pt}\selectfont \cite{icke2020performance}};

% Egypt
\coordinate (Egypt) at ($(Egypt.center)+(.02,.21)$);
\draw[fill=yellow] (Egypt) circle (2.5pt); 
\node[anchor=east] at (Egypt) {\fontsize{10pt}{9.6pt}\selectfont \cite{el2021forecasting}};

% Switzerland
\coordinate (Switzerland) at ($(Switzerland.center)$);
\draw[fill=yellow] (Switzerland) circle (2.5pt); 
\node[anchor=east] at (Switzerland) {\fontsize{10pt}{9.6pt}\selectfont \cite{durrenmatt2012data}};

% Saudi-Arabia
\coordinate (Saudi) at ($(SaudiArabia.center)+(-.37,-.04)$);
\draw[fill=yellow] (Saudi) circle (2.5pt); 
\node[anchor=north] at (Saudi) {\fontsize{10pt}{9.6pt}\selectfont \cite{cheng2020forecasting, dairi2019deep}};

% Jordan
\coordinate (Jordan) at ($(Jordan.center)$);
\draw[fill=yellow] (Jordan) circle (2.5pt); 
\node[anchor=west] at (Jordan) {\fontsize{10pt}{9.6pt}\selectfont \cite{al2024data}};

% Melbourne
\coordinate (Melbourne) at ($(Australia.center)+(.75,-.62)$);
\draw[fill=yellow] (Melbourne) circle (2.5pt); 
\node[anchor=west] at (Melbourne) {\fontsize{10pt}{9.6pt}\selectfont \cite{alali2023unlocking, bagherzadeh2021prediction}};

% Kocaeli, Turkey
\coordinate (Kocaeli) at ($(Turkey.center)+(-.35,.12)$);
\draw[fill=yellow] (Kocaeli) circle (2.5pt); 
\node[anchor=west] at (Kocaeli) {\fontsize{10pt}{9.6pt}\selectfont \cite{ekinci2023application}};

% Boulder, Colorado, US
\coordinate (Boulder) at ($(UnitedStatesOfAmerica.center)+(0.89,-.37)$);
\draw[fill=yellow] (Boulder) circle (2.5pt); 
\node[anchor=north] at (Boulder) {\fontsize{10pt}{9.6pt}\selectfont \cite{harrou2018statistical, newhart2020hybrid}};

% Algeria
\coordinate (Algeria) at ($(Algeria.center)+(0.25,.48)$);
\draw[fill=yellow] (Algeria) circle (2.5pt); 
\node[anchor=north] at (Algeria) {\fontsize{10pt}{9.6pt}\selectfont \cite{mekaoussi2023predicting}};

% Shenzhen and Dongguan, China
\coordinate (Shenzhen) at ($(China.center)+(0.59,-.81)$);
\draw[fill=yellow] (Shenzhen) circle (2.5pt); 
\node[anchor=north] at (Shenzhen) {\fontsize{10pt}{9.6pt}\selectfont \cite{liu2020prediction, ly2022exploring}};

% Doha
\coordinate (Doha) at ($(UnitedArabEmirates.center)+(-0.16,.08)$);
\draw[fill=yellow] (Doha) circle (2.5pt); 
\node[anchor=west] at (Doha) {\fontsize{10pt}{9.6pt}\selectfont \cite{lotfi2019predicting}};

% Nan'an and Chongqing, China
\coordinate (Chongqing) at ($(China.center)+(0.13,-.34)$);
\draw[fill=yellow] (Chongqing) circle (2.5pt); 
\node[anchor=north] at (Chongqing) {\fontsize{10pt}{9.6pt}\selectfont \cite{guo2020data,  wang2023online}};

% Ziyang, China
\coordinate (Ziyang) at ($(China.center)+(0.06,-.32)$);
\draw[fill=yellow] (Ziyang) circle (2.5pt); 
\node[anchor=east] at (Ziyang) {\fontsize{10pt}{9.6pt}\selectfont \cite{huang2023construction}};

% Beijing, China
\coordinate (Ziyang) at ($(China.center)+(0.75,.29)$);
\draw[fill=yellow] (Ziyang) circle (2.5pt); 
\node[anchor=south] at (Ziyang) {\fontsize{10pt}{9.6pt}\selectfont \cite{han2018data, sun2017reduction, wang2021soft, wang2022artificial, zhu2022improved}};

% Univeristy of Alabama, US
\coordinate (Alabama) at ($(UnitedStatesOfAmerica.center)+(1.99,-.68)$);
\draw[fill=yellow] (Alabama) circle (2.5pt); 
\node[anchor=west] at (Alabama) {\fontsize{10pt}{9.6pt}\selectfont \cite{granata2017machine}};

% Zhengzhou, China
\coordinate (Zhengzhou) at ($(China.center)+(0.6,-.09)$);
\draw[fill=yellow] (Zhengzhou) circle (2.5pt); 
\node[anchor=west] at (Zhengzhou) {\fontsize{10pt}{9.6pt}\selectfont \cite{lv2024enhancing}};

% Israel
\coordinate (Israel) at ($(Israel.center)$);
\draw[fill=yellow] (Israel) circle (2.5pt); 
\node[anchor=south] at (Israel) {\fontsize{10pt}{9.6pt}\selectfont \cite{farhi2021prediction}};

%% Treviso, Italy
%\coordinate (Treviso) at ($(Italy.center)+(0.0,.23)$);
%\draw[fill=yellow] (Treviso) circle (2.5pt); 
%\node[anchor=west] at (Treviso) {\fontsize{10pt}{9.6pt}\selectfont \cite{mamandipoor2020monitoring}};

% Italy
\coordinate (Italy) at ($(Italy.center)$);
\draw[fill=yellow] (Italy) circle (2.5pt); 
\node[anchor=north] at (Italy) {\fontsize{10pt}{9.6pt}\selectfont \cite{bellamoli2023machine, bernardelli2020real, mamandipoor2020monitoring}};

% Slovenia
\coordinate (Slovenia) at ($(Slovenia.center)$);
\draw[fill=yellow] (Slovenia) circle (2.5pt); 
\node[anchor=west] at (Slovenia) {\fontsize{10pt}{9.6pt}\selectfont \cite{hvala2020design}};

% Louisville, Kentucky, US
\coordinate (Kentucky) at ($(UnitedStatesOfAmerica.center)+(2.06,-.4)$);
\draw[fill=yellow] (Kentucky) circle (2.5pt); 
\node[anchor=south] at (Kentucky) {\fontsize{10pt}{9.6pt}\selectfont \cite{ebrahimi2017temporal}};

% Solingen-Burg, Germany
\coordinate (Solingen) at ($(Germany.center)+(-0.19,-0.0)$);
\draw[fill=yellow] (Solingen) circle (2.5pt); 
\node[anchor=west] at (Solingen) {\fontsize{10pt}{9.6pt}\selectfont \cite{torregrossa2017data, torregrossa2018machine}};

% Rotterdam, the Netherlands
\coordinate (Rotterdam) at ($(Netherlands.center)+(-0.06,-0.0)$);
\draw[fill=yellow] (Rotterdam) circle (2.5pt); 
\node[anchor=east] at (Rotterdam) {\fontsize{10pt}{9.6pt}\selectfont \cite{vasilaki2018relating}};

% Umeå, Sweden
\coordinate (Umea) at ($(Sweden.center)+(0.12,0.1)$);
\draw[fill=yellow] (Umea) circle (2.5pt); 
\node[anchor=west] at (Umea) {\fontsize{10pt}{9.6pt}\selectfont \cite{wang2022towards, wang2021machine}};

% Shanghai, China
\coordinate (Shanghai) at ($(China.center)+(1.,-.29)$);
\draw[fill=yellow] (Shanghai) circle (2.5pt); 
\node[anchor=west] at (Shanghai) {\fontsize{10pt}{9.6pt}\selectfont \cite{sheng2023exploring, wu2023coupling}};

% Jiangsu Province, China
\coordinate (Jiangsu) at ($(China.center)+(.94,-.27)$);
\draw[fill=yellow] (Jiangsu) circle (2.5pt); 
\node[anchor=east] at (Jiangsu) {\fontsize{10pt}{9.6pt}\selectfont \cite{xie2024hybrid, xu2021integrated}};

% Jiangxi Province, China
\coordinate (Jiangxi) at ($(China.center)+(.75,-.47)$);
\draw[fill=yellow] (Jiangxi) circle (2.5pt); 
\node[anchor=west] at (Jiangxi) {\fontsize{10pt}{9.6pt}\selectfont \cite{yang2022prediction}};

% Japan
\coordinate (Japan) at ($(Japan.center)+(.17,-.07)$);
\draw[fill=yellow] (Japan) circle (2.5pt); 
\node[anchor=south] at (Japan) {\fontsize{10pt}{9.6pt}\selectfont \cite{zhao2022machine}};

% South Korea
\coordinate (Korea) at ($(SouthKorea.center)$);
\draw[fill=yellow] (Korea) circle (2.5pt); 
\node[anchor=west] at (Korea) {\fontsize{10pt}{9.6pt}\selectfont \cite{geng2024multi, guo2015prediction, kim2021machine, yaqub2020modeling, yaqub2022modeling}};

% Brazil
\coordinate (Brazil) at ($(Brazil.center)+(.43,-.42)$);
\draw[fill=yellow] (Brazil) circle (2.5pt); 
\node[anchor=west] at (Brazil) {\fontsize{10pt}{9.6pt}\selectfont \cite{oliveira2002simulation}};

\end{tikzpicture}
}
\caption{The location of the WWTPs providing data for the studies cited in Appendix \ref{sec:literature}, in the cases where they are identified.}
\label{fig:map}
\end{figure}

One of the key challenges in applying machine learning to wastewater treatment lies in the lack of standardization across treatment facilities. Wastewater treatment plants vary significantly, not only in type but also in facility size, process configuration and architecture, equipment, sensors, chemical usage, and other factors. Appendix \ref{sec:literature} summarizes recent literature on machine learning of wastewater treatment, with an emphasis on papers that use real-world data from specific plants. The locations of these plants are shown in Figure \ref{fig:map}.

The map in Figure \ref{fig:map} clearly highlights that the vast majority of studies rely on data from WWTPs located in regions with more stable temperatures than what the Veas plant experiences. Notable exceptions are the papers by Wang et al.\ that use data from a plant in Umeå in northeast Sweden \cite{wang2021machine, wang2022towards}. This leads to one of two main motivations for our study: More research is needed on machine learning of wastewater treatment in colder temperate climates. This need comes from two sides, the process and the modelling: i) the wastewater treatment process in Norway is inherently different from that in regions with more stable and hotter climate, necessitating a specialized approach; ii) big seasonal variations lead to highly non-stationary system dynamics, complicating data-driven modelling approaches.

The other main motivation is to reach a deeper understanding of what is needed to do efficient machine learning of a WWTP. We deem much of the recent literature, summarized in Appendix \ref{sec:literature}, to be somewhat rushed; data originally obtained for manual control and operational oversight is fed into one or more established machine learning methods, and the resulting predictions are compared and measured by their accuracy. While these studies have value, not least in giving some sense of the capabilities of current ML methods, we take a different approach here: the goal of the machine learning performed is not primarily to get accurate predictions at this point, but to take initial steps towards identifying what critical process parameters must be monitored, what quantity and quality the data must have, how the data should be structured, and what characteristics and properties the ML methods should have.

For transparency, and to allow others to build on our work, we publicly share the code for our models and analysis, as well as the data obtained from the pilot reactor at Veas. The code is available on the GitHub repository \url{https://github.com/sintef/veas-denit-pilot} and has been archived at \cite{veas-denit-pilot-code}, while the dataset can be found at \cite{veas-denit-pilot-data}.

\section{Data}

\subsection{Data selection}
The Veas plant has many sensors providing measurements, of which several represent properties that can conceivably have some impact on the denitrification process, directly or indirectly. This study focuses on predicting the nitrate level after denitrification in a pilot reactor, a specific target that distinguishes our work from previous studies which have typically focused on broader nitrogen measures or other wastewater parameters \cite{yaqub2020modeling, yaqub2022modeling, sheng2023exploring, xie2024hybrid}. We hypothesise that the target depends on the following covariates, which we therefore give as input to our models:
\begin{itemize}
    \item \textit{The water temperature:} Measured before sand and grit removal. This is important because temperature significantly affects the rate of microbial activity and the efficiency of biochemical reactions, and is particularly important in climates with high seasonal variability.
    \item \textit{Inlet nitrate nitrogen levels:} Measured in the buffer tank after nitrification in process line no.\ 1, where the water pumped into the pilot reactor comes from. This is used to determine appropriate methanol dosing.
    \item \textit{Inlet oxygen concentration:} Measured in the buffer tank in process line no.\ 1. Monitored because the denitrification process occurs under anoxic conditions, where oxygen must first be respired before nitrogen oxides can be subsequently reduced.
    \item \textit{Inlet ortho-phosphate:} Measured in the buffer tank in process line no.\ 2, which we assume has similar content as the corresponding tank in process line 1, where it is not measured. This is included because it can be a limiting nutrient for microbial growth.
    \item \textit{Turbidity:} I.e., "cloudiness" of water caused by suspended particles. Measured during sedimentation in process line no.\ 1. This is an indicator of particulates entering the reactor, which can cause clogging and impact microbial respiration or fuel fermentative microorganisms.
    \item \textit{Ammonium nitrogen:} Monitored in the buffer tank in process line no.\ 1., as it reflects the performance of the preceding nitrification stage.
    \item \textit{Volumetric methanol dosage:} Methanol is added to the water being pumped to the pilot reactor, as the carbon source for denitrification and oxygen respiration.
    \item \textit{Water flow into the Veas plant}: Measured at the plant inlet. This varies with the time of day and weather conditions. Since it depends on human behaviour, rainfall and melting snow, it is closely connected to the incoming nitrate levels.
    \item \textit{Pressures at the top and bottom of the reactor:} This provides insight into potential clogging of the clay beds, biomass growth, and the possible loss of expanded clay during operation. 
\end{itemize}
For each machine learning model architecture, we perform feature selection among these covariates as part of our optimization procedure, described in Section \ref{sec:model_development}, and a sensitivity analysis to each covariate, presented in Section \ref{sec:nowcasting_fi}.

\subsection{Dataset splitting and preprocessing}\label{sec:data}

\begin{figure}
    \centering
    \includegraphics{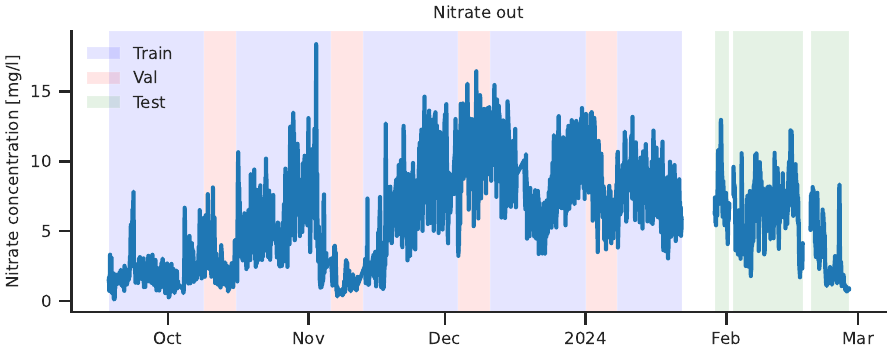}
    \caption{The nitrate out concentration of the Veas pilot, showing one of four cross-validation folds.}
    \label{fig:dataset}
\end{figure}

For both modelling tasks presented below, forecasting and nowcasting, we use a dataset collected over five and a half months from the Veas pilot, with a sampling frequency of 10 minutes. We observe that the data exhibits vastly different characteristics throughout this period. We reserved the last 20\% of the data as a test set, on which we report the results and conduct the analysis presented in the next sections. For the first 80\% of the data, we operate with different splits between training and validation data for i) hyperparameter optimization and choice of the optimal model, and ii) training of the chosen model. For the former, we divided the data into three weeks of training data followed by one week of validation data and so on; see Figure \ref{fig:dataset}. We create in total four cross-validation folds by moving the start of the validation weeks. Each of these constitutes a roughly 60\%, 20\%, 20\% split between the training, validation, and test sets, respectively. For the training of each model after hyperparameter tuning, we use the first 72\% as training data and the subsequent 8\% as validation data.

We note that there is a gap in time before the onset of the test set (and twice within the test set), caused by downtime in the pilot reactor and subsequent lapse of measurements. Data is standardized based on the training dataset before being fed to the models, but we report metric scores based on the original non-transformed data.

The pilot undergoes cleaning typically once every 24 hours, lasting for about an hour. However, there is no strict schedule for when and how long this cleaning is performed. During this time, the sensors will keep producing measurements that are meaningless. To combat this, we detect from the pressure when cleaning is occurring, mask out these readings, and then fill them in based on linear interpolation between the last and first values before and after cleaning, respectively. We experimented with various approaches to dealing with these periods, including no corrective action, linear interpolation of all inputs ($X$) and the output ($y$), and removing this data entirely (leaving gaps in time), and found linear interpolation of only the target variable $y$ (while leaving $X$ as is) to yield best results in terms of model performance.

\section{Methods}\label{sec:mlmodels}
\subsection{Machine learning methods}

We investigate different model architectures of varying complexity. Our comprehensive approach compares multiple machine learning methods, ranging from simple linear models to complex neural networks, on the same dataset.  The methods considered are:
\begin{itemize}
    \item The linear regression method \textit{ElasticNet}, that utilizes both L1 and L2 regularization to mitigate overfitting. ElasticNet has the benefit of being the simplest and most interpretable model, at the cost of low expressiveness (representational power) as it can only learn functions that are linear in its inputs.
    \item On the other side of the expressiveness spectrum we investigate recurrent neural networks (specifically long short-term memory (\textit{LSTM}) networks), and temporal convolutional networks (\textit{TCN}), which can learn to approximate any arbitrary function of the inputs, at the cost of high computational complexity and low interpretability. 
    \item The third major class of models we experiment with is gradient-boosted decision tree methods, namely \textit{XGBoost}. These methods lie somewhere between linear regression and neural networks in expressiveness and interpretability, and typically excel in low data regimes.
\end{itemize}
This comparative analysis is not common in the literature and provides valuable insights into the relative performance of these methods for our specific prediction task, as well as their data requirements and response to various parameters such as the length of input history and selections of covariates.

ElasticNet was tested for modelling sludge production in \cite{ekinci2023application}, while other linear regression models were shown to perform well against nonlinear models for nitrogen prediction in \cite{kim2021machine}. LSTM was proposed for predicting ammonium, total nitrogen and total phosphorous in \cite{yaqub2020modeling}, and XGBoost was proposed for the same model targets in \cite{yaqub2022modeling}. TCN was used for predicting nitrogen concentrations in \cite{sheng2023exploring}, while a combination of LSTM and TCN was proposed for this task in \cite{xie2024hybrid}. XGBoost was shown to outperform random forest models for predicting suspended solids and phosphate in \cite{wang2022towards}.

We view the utility of the data-driven modelling through two different lenses in this paper: In the first, we wish to use the model to learn about the process and how the concentration of nitrate out of the denitrification reactor is affected by the other measured covariates. This is useful for increasing our understanding of the process, and for the development of a digital twin of the full treatment plant. To this end, we train the models to \textit{nowcast} the concentration of nitrate in the output of the reactor using only the other covariates as inputs. Secondly, given previous and current measurements, we train the models to forecast future concentrations of nitrate, which can be employed for automatic control or as decision-making support for operators. These cases are detailed in the following two subsections.

\subsection{Nowcasting nitrate concentration}
In this case, we wish to learn how the various measured quantities affects the denitrification process, and how they interact with each other. A reasonable estimate of the output concentration of nitrate at time $t$ is that it should be similar to the concentration at time $t-1$. Thus, in order to avoid having the model converge to this simple hypothesis, we do not provide historical measurements of the output concentration of nitrate to the model. Instead, we train the models to infer the output concentration based on current and historical measurements of the other covariates, i.e.\ nowcasting:

\begin{align}
    argmin_\theta \enspace &L(f_\theta(X_{t-h:t}), y_t) \label{eq:nowcasting_objective} \\
    X_{t-h:t} = &\begin{bmatrix}
                x^1_{t} & \dots & x^n_{t} \\
                \dots & \dots & \dots \\
                x^1_{t-h} & \dots & x^n_{t-h}
             \end{bmatrix}
\end{align} 
where $X_{t-h:t}$ is the collection of $h$-length time-series measurements of $n$ covariates $\{x^1, \dots, x^n\}$, $f_\theta$ is the model with learnable parameters $\theta$, and $L$ is some objective function giving a measure on the difference between the model output and the concentration of nitrate in the output of the reactor, $y$. In this case study, we use the mean squared error (MSE) as $L$. The learnable parameters $\theta$ are adjusted to minimize the objective $L$ on the training data, and we further optimize hyperparameters of the model architectures using cross-validation data, as detailed in sections \ref{sec:data} and \ref{sec:model_development}.

\subsection{Forecasting future nitrate concentration}
When forecasting future behaviour of the process, we do not care if the model relies heavily on the history of the output concentration. Instead, we only care about its efficacy in predicting the future concentrations of nitrate. Therefore, when forecasting, we include historic measurements of nitrate as inputs to the model and train it to predict future values:

\begin{align}\label{eq:forecasting_objective}
    argmin_\theta \enspace &L(f_\theta(X_{t-h:t}, y_{t-h:t}), y_{t+1})
\end{align}

Note that the models are trained on one-step predictions like in \eqref{eq:forecasting_objective}, but when we evaluate on the validation and test sets we predict six steps ahead, which corresponds to one hour into the future.

\subsection{Model development methodology}\label{sec:model_development}
For each model architecture, we identify a set of hyperparameters that we can adjust to optimize the performance of the models. These are specific to each model architecture and include parameters like the size of the neural networks and the size of the trees in XGBoost, the strength of regularization objectives to limit overfitting, etc., as well as some hyperparameters that are common to all models such as which input covariates to give the models and how much history is included (the $X$ and $h$ parameters in \eqref{eq:nowcasting_objective}). We use hyperparameter optimization to determine the optimal set of parameters for each architecture that perform on average the best over the four cross-validation folds, independently for the two tasks forecasting and nowcasting. Then, with these optimized hyperparameters, we train one model on the training and validation datasets, with the 72/8 split explained in Section \ref{sec:data}, where the validation data is used for early-stopping. Finally, these models are then deployed on the test set. The resulting predictions are presented and analyzed in the next sections.

With these optimized models we investigate the effects of varying the $h$ parameter controlling how much history the models consider, as well as what combination of the covariates $x$ that yields the best models. Note that while we already consider these parameters in the hyperparameter optimization procedure described in the preceding paragraph, they are in that case selected based on the cross-validation data, while the analysis we present for $x$ and $h$ in Section \ref{sec:nowcasting_fi} is based on the test data, i.e.\ a posteriori information that is not available at the time of model development.

\section{Results and discussion}\label{sec:results}

\begin{figure}
    \centering
    \includegraphics{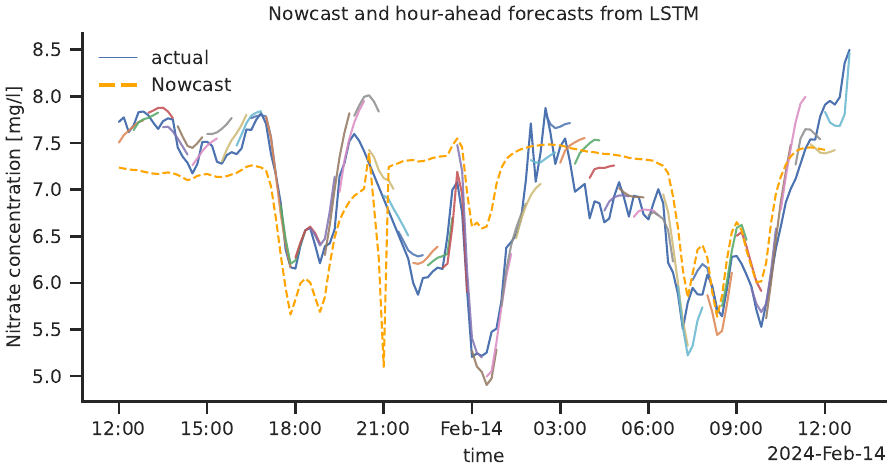}
    \caption{Example of nowcast and forecasts on test set from the LSTM model. Each forecast, given by a colored line, is from 10 minutes to one hour ahead.}
    \label{fig:forecast_example}
\end{figure}

\begin{figure}
    \centering
    \begin{subfigure}[b]{0.49\textwidth}
        \centering
        \includegraphics{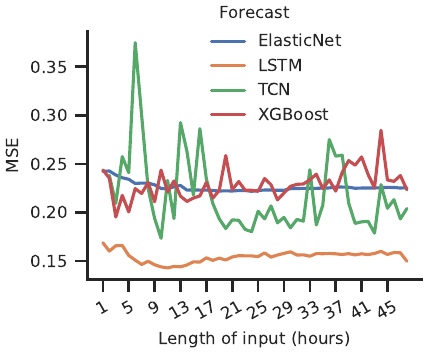}
        \caption{Forecasting}
        \label{fig:lags:forecast}
    \end{subfigure}
    \begin{subfigure}[b]{0.49\textwidth}
        \centering
        \includegraphics{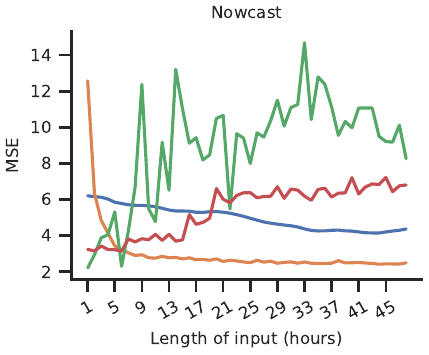}
        \caption{Nowcasting}
        \label{fig:lags:nowcast}
    \end{subfigure}
    \caption{The effect of varying the input length to the models after hyperparameter optimization on the test set on the MSE metric.}
    \label{fig:lags}
\end{figure}

Table \ref{tab:test_set_metrics} shows the results of applying the optimized models to the test set, along with several baselines for comparison. For the methods with a stochastic training process (i.e.\ LSTM, TCN, XGBoost) we train 10 models with different seeds and report mean and standard deviation over those seeds. We can immediately observe that there is a considerable discrepancy between the training and validation data on one hand, and the test data on the other. That is, these models are optimized to perform well on the cross validation data (and exhibit good performance on the validation data MSE in Table \ref{tab:test_set_metrics}), but perform much worse on the test set, sometimes even worse than the baselines. Furthermore, by comparing Figures \ref{fig:lags} and \ref{fig:inputs} with the hyperparameters identified as optimal with respect to the validation data (see model configuration files in the project code base), we see that changing the $h$ and $X$ parameters can drastically improve the performance on the test data. This is a general challenge with data-driven modelling, where the available training data is not representative for the desired model task in operation. The ranking of the models is the same for both tasks: LSTM performs the best, followed by ElasticNet, XGBoost, and finally TCN. TCN performs well on the validation set for both tasks but has a significant increase in metric scores on the test set, indicating overfitting. On the other hand, ElasticNet performs the worst of all the models on the validation data, but the simpler linear functions it learns transfers better to the test set. Examples of the forecasts and nowcasts produced by the LSTM model are illustrated in Figure \ref{fig:forecast_example}.

The baselines we have used are: 1) TrainingMean, which always predicts the mean value of the training set, 2) RunningMean, which computes a running mean of the values seen so far in the current dataset (note that this is a tough baseline to beat that is using information not available to the models), 3) Seasonal, which predicts that the next datapoint will be the same as the preceding datapoint, or in the case of forecasting that the next hour will be the same as the preceding hour, i.e.\ a time-lagged version of the output, and 4) TrendN, that fits a line between the preceding N datapoints which it uses to extrapolate ahead in time.

\begin{table}
    \centering
    \caption{Results for the various model architectures on the two tasks, nowcasting and forecasting, sorted by test set MSE. Note that metrics are calculated on the original non-transformed data.}
    \begin{tabular}{llllllllll}%llcccccccc
        \toprule
        \multirow{2}{*}{Task} & \multirow{2}{*}{Model} & Val & \multicolumn{2}{c}{Test} \\
        \cmidrule(lr){4-5}
        & & MSE & MSE $\uparrow$ & MAE \\
        \midrule
        \multirow{6}{*}{Nowcast}
        & LSTM & 3.98 $\pm$ 0.54 & 2.82 $\pm$ 0.19 & 1.43 $\pm$ 0.05 \\
        & ElasticNet & 4.18 & 4.25 & 1.43 \\
        & BaselineTestRunningMean & 3.79 & 4.42 & 1.85 \\
        & XGBoost & 2.67 $\pm$ 0.03 & 6.17 $\pm$ 0.45 & 1.90 $\pm$ 0.05 \\
        & TCN & 1.96 $\pm$ 0.19 & 6.80 $\pm$ 1.26 & 1.99 $\pm$ 0.17 \\
        & BaselineTrainingMean & 8.18 & 7.18 & 2.09 \\
        \midrule
        \multirow{8}{*}{Forecast}
        & LSTM & 0.19 $\pm$ 0.01 & 0.15 $\pm$ 0.02 & 0.25 $\pm$ 0.01 \\
        & ElasticNet & 0.32 & 0.23 & 0.32 \\
        & XGBoost & 0.31 $\pm$ 0.01 & 0.23 $\pm$ 0.01 & 0.32 $\pm$ 0.01 \\
        & TCN & 0.21 $\pm$ 0.01 & 0.24 $\pm$ 0.03 & 0.31 $\pm$ 0.01 \\
        & BaselineTrend6 & 0.75 & 0.50 & 0.43 \\
        & BaselineSeasonal & 0.94 & 0.51 & 0.49 \\
        & BaselineTrend3 & 1.07 & 0.73 & 0.50 \\
        & BaselineTrainingMean & 37.05 & 7.16 & 2.08 \\
        \bottomrule
    \end{tabular}
    \label{tab:test_set_metrics}
\end{table}

\begin{figure}
    \centering
    \includegraphics{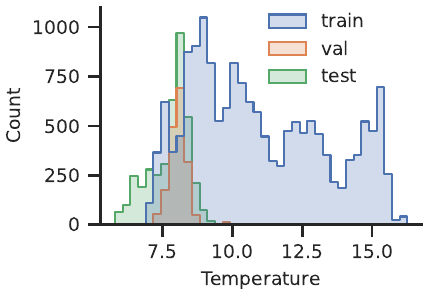}
    \caption{The distribution of the temperature variable across the three datasets.}
    \label{fig:temp_distribution}
\end{figure}

\begin{figure}
    \centering
    \includegraphics{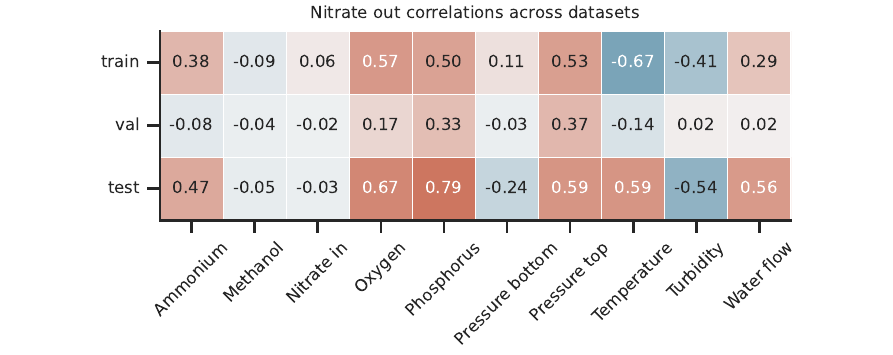}
    \caption{Feature correlations for the three dataset splits.}
    \label{fig:feature_correlations}
\end{figure}

\subsection{Nowcasting covariate selection}\label{sec:nowcasting_fi}

\begin{figure}
    \centering
    \begin{subfigure}[b]{\textwidth}
        \centering
        \includegraphics{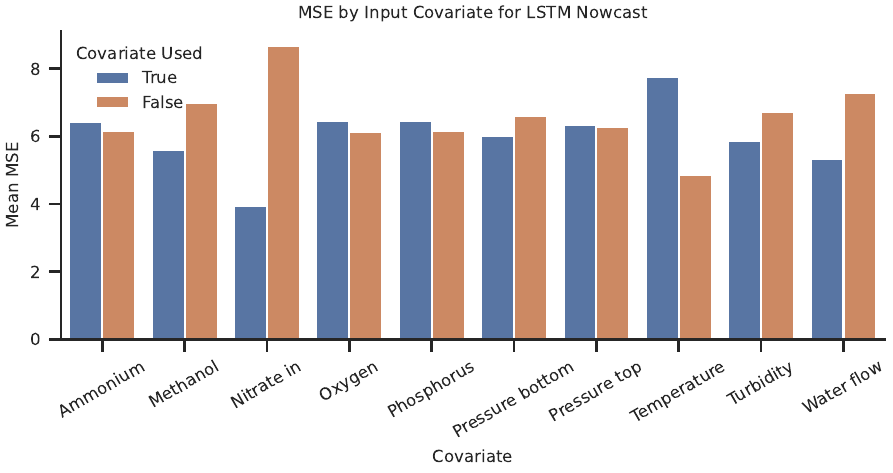}
        \label{fig:inputs:nowcast:rnn:onoff}
    \end{subfigure}
    
    \begin{subfigure}[b]{\textwidth}
        \centering
        \includegraphics{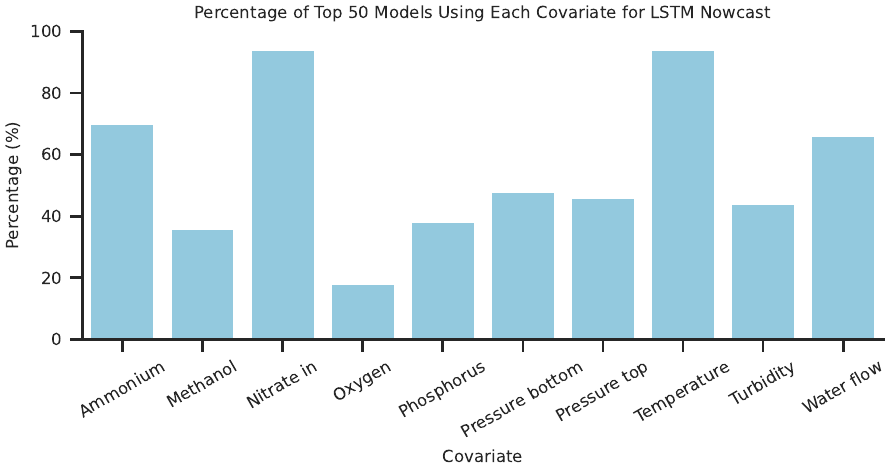}
        \label{fig:inputs:nowcast:rnn:top}
    \end{subfigure}
    \caption{Importance of each covariate to the LSTM method. We construct all 1024 combinations of the 10 covariates, and measure the mean MSE by all models that include a given covariate vs all models that do not include the covariate (top), and the percentage of the top 50 models of the 1024 that include a given covariate (bottom).}
    \label{fig:inputs}
\end{figure}

Figure \ref{fig:inputs} shows the result of the input covariate analysis on the test set for the LSTM method, the rest are found in Appendix \ref{sec:app:plots}. Here we have created all 1024 combinations of the 10 input covariates $x$, and trained a model for each of these combinations and for each model architecture. The top plots show the mean MSE on the test set for all models that include a given covariate vs the mean MSE of all models that do not include that covariate, giving an indication of the average contribution of that covariate to the models. A large discrepancy between having and not having a given covariate in these plots therefore indicate that this covariate provides useful information, or conversely, that the information it provides differs significantly between the training and test set, and as such teaches the model spurious relationships that are destructive to its performance on the test set. A small discrepancy indicates that the models have not learned to correlate the covariate significantly with the output. The bottom plots show the prevalence of each covariate among the top 50 (~5\%) of the models, and therefore demonstrates the maximum potential unlocked by each covariate as a supplement to its average contribution as previously discussed.

These results show that the covariates nitrate in, water flow and methanol are generally the most positively impactful covariates across the different model architectures. Nitrate in stands out as especially impactful. This is also the covariate occurring in the most top models, together with temperature. However, temperature consistently shows a marked negative impact across all model architectures. Inspecting the distribution of the temperature variable for the training set compared to the test set shown in Figure \ref{fig:temp_distribution}, we can see a clear difference, which follows from the training set stemming from autumn going into the winter season, and the test set stemming from February, yielding a 4$\degree$ degree difference in the mean temperature of the two datasets.

\subsection{Identification and analysis of anomalous predictions}
Once the models have been trained and tested, a detailed study and analysis of the resulting predictions can offer valuable insight -- both in the models and the process they represent. For this purpose, nowcasting models are best suited. Since the wastewater process is highly non-stationary, a visual inspection of the predicted nitrate over time against the actual measurements can give information on its performance, beyond the error scores shown in Table \ref{tab:test_set_metrics}. By identifying characteristic patterns in periods where the model fails to give an accurate prediction, we can better understand which sensor data and machine learning methodologies are needed for improved modelling in the future.

For the type of time-dependent process we are modelling here, we have identified three classes of anomalous predictions that we aim to identify:
\begin{itemize}
    \item[1.] There is a notable peak in the target that the model fails to predicts.
    \item[2.] The model predicts a significant peak that does not occur in the target data.
    \item[3.] The model predicts the variations in the target accurately but has a consistent under- or over-estimation bias, before readjusting.
\end{itemize}
Identification of these periods can be made on either the training, validation or test set, having different meaning for each. We judge it most purposeful to analyze the validation set, as early stopping will choose the best-performing model on the validation data, and we want to identify when the anomalies above occur during ideal circumstances. Treating different parts of the full data set as the validation set while training on the remaining data, we found several occurrences of the first and third effects above, but none of the second, when evaluating the best-performing model.

\begin{figure}[htb]
    \centering
    \includegraphics{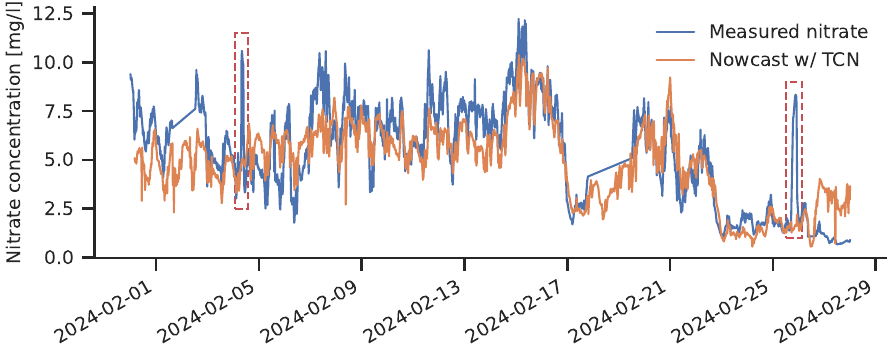}
    \caption{The prediction of the best nowcast model on the validation set, as evaluated by the MSE on the same set. The dashed red rectangles mark two peaks in the measured nitrate not predicted by the model.}
    \label{fig:tcn_nowcast}
\end{figure}

We concentrate here on the last 20\% of the data, corresponding to what was used as the test set in Section \ref{sec:results}. For this analysis, we trained the four different types of models, with the hyperparameters identified by the aforementioned optimization method, and saved each model after each training epoch. Then we picked the model with the lowest MSE on the validation set. This was the TCN model after only one epoch of training; see Figure \ref{fig:tcn_nowcast}. This greatly outperformed the other models, including the TCN model that performed best on the validation set used for the TCN model reported on in Table \ref{tab:test_set_metrics}. This reaffirms the points made at the beginning of Section \ref{sec:results} about the validation and test set representing vastly different dynamics, which is also showing its effect in Figure \ref{fig:lags}.

\begin{figure}
    \centering
    \begin{subfigure}[b]{0.475\textwidth}
        \centering
        \includegraphics[width=\textwidth]{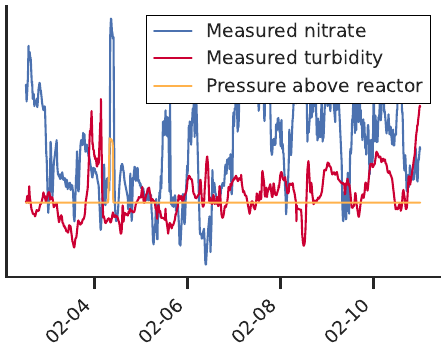}
    \end{subfigure}\hspace{18pt}
    \begin{subfigure}[b]{0.475\textwidth}
        \centering
        \includegraphics[width=\textwidth]{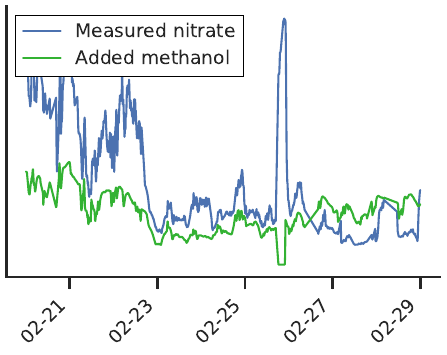}
    \end{subfigure}
    \caption{Measured nitrate and covariates that we hypothesize cause the peaks identified in Figure \ref{fig:tcn_nowcast}.}
    \label{fig:reasons}
\end{figure}

The reason for the second peak not being detected is quickly identified: methanol is not added to the reactor for four hours, due to a fault in the dosing process; see the right plot in Figure \ref{fig:reasons}. Even though the same fault occurred thrice in the training data, the model has not learned the strong connection between methanol dosing and nitrate levels after denitrification. We deem that this is due to the lack of variation in methanol dosing for the large majority of the training data, where methanol has been dosed according to a formula largely dependent on the incoming nitrate levels. Therefore, modifying the model to detect deviations from this formula could improve its ability to predict peaks, such as the second one in \ref{fig:tcn_nowcast}, and enhance its overall performance.

The first peak is more difficult to explain. Analysing all the input data, we find two abnormalities, shown in the left plot in Figure \ref{fig:reasons}: For one, there is a significant temporary increase in turbidity in the incoming water between 12 and four hours before the start of the nitrate peak. Second, a spike in the pressure at the top of the reactor coincides with this peak; under normal conditions, this pressure should be stable. We hypothesise that deviations in the pressure measurements result from expanded clay floating to the surface, an issue discussed further in Section \ref{sec:leca}. While established theories do not offer a clear explanation for how and why these abnormalities affect the nitrate measurement, these observations provide a basis for developing new hypotheses that warrant further investigation into the pilot system and the treatment process.

Although both identified cases of anomalous predictions fall within the same category, they arise from different causes and offer distinct insights. The first case is crucial for refining the model, while the second has the potential to reveal new insights into the process itself, which can subsequently inform model improvements in the next stage.

\subsection{Data limitations}
Despite the extensive efforts in developing optimal models through hyperparameter optimization, we observe that these optimal models on the training and validation data perform significantly worse on the test data (Table \ref{tab:test_set_metrics}), and that other parameters perform better on the test data (Figure \ref{fig:lags}, \ref{fig:inputs}). This is in large part due to lacking data quality and data quantity. The analysis presented in Section \ref{sec:results} highlights that the denitrification process is heavily seasonal, which is especially evident in the effect of the temperature variable. Temperature is a rate-determining factor in biological systems, and low temperature increases the solubility of oxygen. The rate of diffusion will also increase with temperature (which can be important in a biofilm) -- and which, in principle, may lead to increased methanol dosing as a higher concentration gradient will also increase diffusion rates. The test set period in February is around the typical time of minimum wastewater temperature in northern climates due to the melting of snow. As such, multiple years' worth of data is required in order to have several realizations of these seasonalities demonstrated in the data to learn from. This highlights the importance of more research on WWTPs in northern climates that face unique challenges not typically addressed in the literature dominated by more centrally located facilities, as shown in Figure \ref{fig:map}.
The fact that the nonlinear models perform best on the training and validation data but transfer worse to the test data than the linear model suggests either consistent overfitting or that there are nonlinear relationships to be learned in the process. Both of these outcomes are caused by a lack of data quality (unmeasured covariates, noisy measurements) and data quantity (insufficient number of realizations of patterns to generalize from). This is supported by the broader time series forecasting literature, where simple linear models have been shown to generalize better than more complex nonlinear models in low-data or distribution-shift settings \cite{zeng2023transformers, chen2023tsmixer}.

\subsection{Loss of expanded clay}\label{sec:leca}
A further complicating factor that we discovered throughout this work is the effect of loss of biofilm carrier material (i.e.\ expanded clay) in the denitrification reactor. The expanded clay pellets harbour denitrifying bacteria, meaning that the amount of expanded clay directly affects the nitrate reduction rate. We observed that expanded clay intermittently floats to the surface of the reactor and is subsequently lost through the outlet, negatively impacting the denitrification capacity. Although this issue also occurs in the main process, it is effectively mitigated by physical barriers, making it a non-issue there. As a result, the problem was largely overlooked in the pilot process until it was highlighted during inspection of data.

\begin{figure}[h!]
    \centering
    \includegraphics{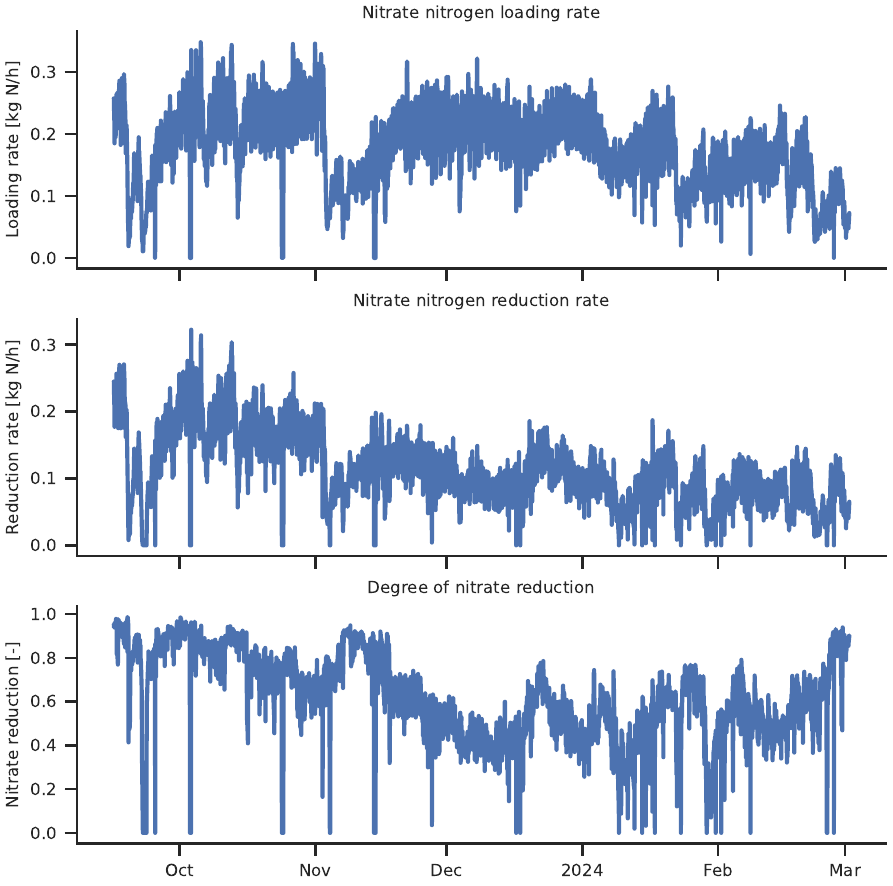}
    \caption{The input concentration of nitrate to the reactor, and the degree of nitrate reduction.}
    \label{fig:nitrate}
\end{figure}

The loss of expanded clay is a significant obstacle for accurate machine learning, since the amount of expanded clay is not measured systematically. However, manual measurements were taken to confirm the substantial loss of expanded clay: The amount of pellets was measured to be 3 $m^3$ at the onset of the dataset, before not being measured again until the end of February when it was down to 1 $m^3$. This discrepancy will significantly impact the nitrate conversion and the dynamics of the process. Information about the amount of expanded clay has not been given as input to the models, and is one factor explaining the difficulties in generalizing patterns learned in the training data to the test data period. The biofilm carrier material was refilled at the start of March, and this clearly improves the nitrate conversion rate, as seen at the end of the plot in Figure \ref{fig:nitrate}; however, note that this period is not included in the dataset of this paper.

We conducted tests where the estimated volume of expanded clay was included as input to the models, but this did not result in a significant improvement to the predictions. This was as expected, as the estimates were imprecise, derived from only two manual measurements and disturbances in the pressure readings, which we hypothesize are related to loss of expanded clay. We plan for future work on modelling of this pilot to rely on frequent measurements of the volume of biofilm carrier material.

\subsection{Prediction target}
Crucial information may be missing because Veas does not measure some important parameters. Of particular concern is the assumption that the change in nitrate concentration across the reactor is equivalent to the complete conversion of nitrogen to molecular nitrogen (i.e., complete denitrification). However, the denitrification process is sequential, with nitrate being reduced through the intermediates nitrite, nitric oxide, and nitrous oxide. The accumulation of these intermediates can indicate an imbalance in the process. Some intermediates, like nitrite ($\mathrm{NO_2^-}$) and nitric oxide ($\ce{NO}$), can be toxic to bacteria \cite{pan2015evaluating}, and is overlooked due to the lack of measurements. Nitrous oxide ($\ce{N_2O}$), while not toxic, is a potent greenhouse gas that can significantly impact the plant’s overall greenhouse gas footprint \cite{kampschreur2009nitrous}. Additionally, the nitrate sensors measure the combined concentration of both nitrate and nitrite nitrogen. High nitrite concentrations at the inlet, recorded as nitrate, may indicate an unstable nitrification process upstream, potentially leading to toxic levels of nitrite being fed into the pilot system.

In this work we chose to model the denitrification process by predicting the output concentration of nitrate. One could instead predict the conversion rate of input nitrate to output nitrate, which may have made the analysis of the results more straightforward. 
Further, we have fed the covariates to the models as raw time-series, deferring the responsibility of feature engineering to the models. If the output concentration of nitrate is a complex nonlinear function of the input covariates, manually performing feature engineering could improve performance, especially for the models that have no (linear regression) or little (XGBoost) ability to construct complex features from their inputs.

\section{Conclusions}

This paper has presented a systematic approach to modelling of a pilot WWTP denitrification reactor, and identified key challenges concerning the data requirements to develop optimal models for use in a digital twin of the process. We found that there were great differences in the data distributions and input-output behaviour over time due to three main factors: 1) Unmeasured factors affecting the process such as the amount of biofilm carrier material, 2) seasonalities in the process largely due to temperature variations and reduced influent loads during holiday periods, and 3) lack of exploration in process parameter combinations, as evidenced by the inability of the models to learn the direct relationship between lack of dosed methanol and corresponding increase in output nitrate (Figure \ref{fig:tcn_nowcast}), as methanol is in the training data always dosed according to a single strategy. Finally, the relative success of linear models compared to nonlinear models and the fact that these transfer better from older data to later data suggests that higher quantities and more informative data is necessary to model WWTPs with high accuracy. The analysis of covariate inputs revealed that some covariates were consistently more impactful across the different model architectures, in particular nitrate in concentration, dosed methanol, and water flow into the plant. On the other hand, the temperature variable was found to have a marked negative impact due to the aforementioned data challenges. Based on these findings and the extensive literature review we have presented, we call for more research on modelling of WWTPs in colder temperate climates that face unique challenges. The insights presented will be used as a foundation for further instrumentation of the main process at Veas and for designing models where the ultimate goal is to employ them in closed-loop control.

\subsection*{CRediT authorship contribution statement}
\textbf{Eivind Bøhn:} Writing – original draft, Writing – review and editing, Conceptualization, Data curation, Formal analysis, Methodology, Software, Visualization.
\textbf{Sølve Eidnes:} Writing – original draft, Writing – review and editing, Conceptualization, Data curation, Formal analysis.
\textbf{Kjell Rune Jonassen:} Writing – original draft, Writing – review and editing, Conceptualization, Investigation.

\subsection*{Declaration of generative AI in the writing process}
During the preparation of this work, the authors used ChatGPT to get suggestions on how to improve the readability of certain parts of the manuscript. The authors reviewed, edited, and reformulated the text independently and take full responsibility for the content of the published article

\subsection*{Declaration of competing interest}
The authors declare that they have no known competing financial interests or personal relationships that could have appeared to influence the work reported in this paper.

\subsection*{Acknowledgements}
This work is funded by the European Union under grant agreement
no.\ 101135932. Views and opinions expressed are however those of the authors only and do not necessarily reflect those of the European Union or Directorate-General for Communications Networks, Content and Technology. Neither the European Union nor the granting authority can be held responsible for them.

\bibliographystyle{unsrt} % numeric citations, in order of appearance
\bibliography{veaspilot}

\appendix

\section{The wastewater treatment process at Veas}\label{sec:veas_detailed}

The Veas WWTP exemplifies the variation among treatment facilities with its unique operational characteristics. The plant employs a combination of mechanical, chemical, and biological treatment methods, notable for its efficient mixing of precipitation chemicals, effective solids separation, and an exceptionally short total water retention time of less than 3 hours. This section outlines the key treatment steps, with a focus on the fixed bed biological nitrogen removal processes employed at Veas, and the pilot denitrification system used in this study.

Figure \ref{fig:veas} (main paper) presents a simplified flow chart of the process at Veas. The wastewater is screen filtered before being directed to aerated grit chambers, where sand, smaller particles, and fats are removed. Chemical treatment is initiated in these chambers with the addition of aluminium chloride and ferric chloride, which precipitate phosphorus and initiate micro-flocculation, i.e.\ bridging of particles due to the formation of hydroxide bonds. The dosing of precipitation chemicals is controlled to give a low concentration of soluble phosphorus, and to maintain sufficient buffer capacity. This buffer supports activity and growth of bacteria in the downstream biological treatment. Before the wastewater enters the sedimentation basins, a charged polyacrylamide polymer is added to aggregate the micro-flocs, which settle at the bottom along with insoluble phosphate salts. The sludge is collected from the bottom of the basins. It is further processed into biogas, ammonium sulphate, and organic fertilizer, whilst the particle-free water phase, containing dissolved nitrogen (ammonium) and dissolved carbon, is directed to the biological treatment (section \ref{sec:bio}). Solids retrieved after sedimentation, in the form of wastewater sludge, is processed into biogas, ammonium sulphate and soil enrichment products; this is omitted from Figure \ref{fig:veas}.

%\begin{figure}[ht!]
%    \centering
%    \resizebox{\textwidth}{!}{\input{veastikz_alt}}
%\caption{\textit{Alternative figure, if we do not want to include backwashing lines}.}
%\label{fig:veas}
%\end{figure}

\subsection{Biological treatment} \label{sec:bio}

The biological treatment at Veas consists of a two-step process: nitrification followed by denitrification. Both processes are conducted in stationary fixed-bed biofilm reactors that utilize expanded clay, with a size in the range of 3-5 mm, as the biofilm carrier material. This expanded clay has a density just above that of water, and its porosity results in a high catalytic capacity, meaning the total surface area per volume for biofilm growth. Each of the eight process lines is equipped with four reactors dedicated to nitrification, and four reactors for denitrification. Wastewater enters from the bottom of the reactors and flows through the expanded clay bed, before being discharged from the top.

The nitrification reactors cover an area of $80$ $\mathrm{ m}^2$ and are filled with four meters of expanded clay. The denitrification reactors are slightly smaller, with an area of $66$ $\mathrm{ m}^2$ and a depth of three meters of expanded clay. Being fixed-bed reactors they serve a dual purpose: they facilitate the biological conversion of dissolved nitrogen and act as physical filters by retaining particles that were not separated during sedimentation. This design eliminates the need for any final clarification of the effluent water, thereby optimizing space within the plant. The average contact time between the wastewater and the biofilm carrier is 16 minutes in the nitrification reactors and 8 minutes in the denitrification reactors. The biofilm carriers are backwashed periodically to remove particulates and excess biomass; the nitrification reactors every 12 hours and the denitrification reactors every 22 hours. In this backwashing process treated wastewater and air enters from the bottom of the reactors at high velocities, generating shear forces that "liquefy" the clay bed. Particles are released, and excess biomass is removed due to collisions between the clay particles. Particles and biofilm debris are removed by overflow and returned to the inlet of the Veas plant.% The backwashing lines are ommitted from Figure \ref{fig:veas}.

\subsection{Biofilm processes}
A biofilm process, like the ones employed at Veas, involves a community of bacteria that adhere to a surface and form a structured multicellular layer known as a biofilm. Due to the physical structure, biofilm processes are constrained by mass transfer limitations, and substrate gradients thus develop within the biofilm. Only the bacteria residing in the outer layers of the biofilm have direct access to substrates in the water bulk phase, thereby generally having a crucial role in metabolic activity. This dynamic significantly influences both the design and operational strategies of biofilm reactors, as well as the overall microbial ecology within the biofilm structure. The process efficiency can be influenced by a plethora of process variables and process conditions. E.g., temperature will affect conversions kinetics and solubility of gaseous substrates, whilst variations in flow rate, substrate concentrations, and pH can disrupt and alter the microbial communities and influence process efficiency.

\subsubsection{Nitrification}
Nitrification is an aerobic process and involves a two-step sequential oxidation catalysed by two different groups of bacteria\footnote{Ammonia-oxidizing archaea (AOA) and complete ammonia oxidation (comammox) are generally regarded as having a limited contribution to the total nitrification potential in wastewater systems.}. Ammonia oxidizing bacteria (AOB) oxidises ammonium via hydroxylamine to nitrite:
\begin{align*}
\ce{NH_4^+ + 3/2 O_2 -> NO_2^- + H_2O + 2H^+},
\end{align*}
and nitrite oxidising bacteria (NOB) oxidises nitrite to nitrate:
\begin{align*}
\ce{NO_2^- + 1/2 O_2 -> NO_3^-}.
\end{align*}

Both the AOBs and the NOBs are autotrophic, meaning that they meet their carbon requirements from dissolved \ce{CO_2} and energy requirements from the oxidation of ammonia to nitrite and nitrite to nitrate. Effective nitrification requires that most of the organic matter in the wastewater is removed, since elevated carbon levels can promote the proliferation of heterotrophic bacteria, which obtain carbon- and energy requirements from organic compounds. Heterotrophic bacteria will outcompete nitrifying organisms by being more efficient at scavenging oxygen.

AOBs and NOBs are sensitive to pH, because this affects the equilibrium between ammonium and free ammonia (\ce{NH_3}), and nitrite and nitrous acid (\ce{HNO_2}), both of which can inhibit AOB and NOB activity due to cytotoxicity; the molecules can freely transverse over bacterial cell walls and protonate/deprotonate inside the cells, leading to a change in intracellular pH. Ammonia concentrations between 10 and 150 mg N/L and nitrous acid levels between 0.1 and 10 mg N/L have been reported to inhibit AOB and NOB \cite{Anthonisen1976inhibition}. Oxidation of ammonium by AOB generates protons, which lowers pH. Hence, it is important to maintain a sufficiently high buffer capacity in wastewater to keep pH stable around seven.

\subsubsection{Denitrification}
Nitrification is essential for denitrification; without it, biological nitrogen removal cannot occur because nitrification converts ammonia into nitrate, which serves as the substrate for denitrifying bacteria. Denitrifying bacteria are facultative aerobic heterotrophs that, when oxygen concentrations are low, maintain an anaerobic respiration through the step-wise reduction of nitrate (\ce{NO_3^-}) to nitrite (\ce{NO_2^-}), then to nitric oxide (\ce{NO}), nitrous oxide (\ce{N_2O}), and finally to nitrogen gas (\ce{N_2}). Molecular nitrogen (\ce{N_2}) is released to the atmosphere, effectively removing nitrogen from the wastewater. As heterotrophs they require available organic carbon to sustain their metabolism. The reduction can be summarized as
\begin{align*}
\ce{2NO_3^{-} + 10e^{-} +12H^{+} -> N_2 + 6H_2O},
\end{align*}
where electrons are supplied from organic carbon. At Veas, a controlled addition of carbon is supplied as methanol, with a yearly consumption in the range of 4 500 tonnes. The water exiting the aerated nitrification reactors contains a high concentration of dissolved oxygen, which must be metabolized (lowered) before denitrification can occur. This additional oxygen consumption leads to an increased demand for methanol. Methanol is added as the limiting factor for denitrification, as the plants control the dosing by always having some residual nitrite in the effluent. Residual nitrate will also limit the activity of fermentative organisms in the biofilm, which can produce small volatile fatty acids and cause a pH drop, as respiration is favoured over fermentation when electron acceptors are present in excess.

It is fairly common for a portion of the denitrifying community in wastewater systems to have incomplete denitrification pathways, missing one to three of the genes coding for the enzymes involved in the stepwise reduction of \ce{NO_3^-} to \ce{N_2}. As a result of this modularity, organisms that lack the gene coding for i.e.\ \ce{N_2O} reductase (nosZ) become net emitters of \ce{N_2O}, while those with only the nosZ gene act as net sinks for \ce{N_2O}. Additionally, organisms with a fully functional denitrification pathway can either be significant sinks or sources of \ce{N_2O}, depending on their regulatory biology. Low pH also hampers the maturation of a functional \ce{N_2O} reductase enzyme, which can lead to increased \ce{N_2O} emissions.

\subsection{Veas denitrification pilot}\label{sec:pilot}

Veas has established two pilot reactors for denitrification, each with a footprint of $1$ $\mathrm{m^2}$ and, as in the main process, packed with three meters of expanded clay, which acts as a biofilm carrier material. Each pilot has the capacity to treat wastewater corresponding to a load of 800 persons, and receive water from the buffer volume following nitrification in process line no.\ 1; ref.\ Figure \ref{fig:veas}. Washing water for the backwash procedure is also supplied from this buffer volume. One reactor is dedicated to various process optimization experiments, while the second serves as a baseline, with methanol added to replicate conditions similar to the plants process. Since the denitrification pilot is used for experimental purposes, the outlet water is pumped back to the beginning of the treatment process, rather than being disposed into the Oslofjord.

The methanol dosage is controlled based on measurements from the storage tank, where the flow of methanol to the reactor, in $\mathrm{mg}/\mathrm{l}$ and denoted $Q'_\text{m}$, is set by
\begin{align*}
Q'_\text{m} = Q_\text{w}(k_{1}C_{\ce{NO_{3}-N},\text{in}} - k_{1} C_{\ce{NO_{3}-N},\text{out}} +k_{2}C_{\ce{O_{2}}})
\end{align*}
where $C_{\ce{NO_{3}-N},\text{in}}$ and $C_{\ce{O_{2}}}$ are the concentrations (in $\mathrm{mg}/\mathrm{l}$) of nitrate nitrogen and oxygen in the incoming wastewater and $C_{\ce{NO_{3}-N},\text{out}}$ is the desired concentration of nitrate nitrogen after denitrification, and $Q_\text{w}$ is the water flow in $\mathrm{l}/\mathrm{s}$. The constant $k_1$ is set by the operators, while $k_2$ is based on how much methanol is consumed when the bacteria respires oxygen and consumes methanol. Unlike in the main  process, nitrogen levels after denitrification are not an input to this control. Consequently, the system operates as an open-loop control process, lacking a feedback mechanism.

\renewcommand{\thetable}{B.\arabic{table}}
\setcounter{table}{0}
\section{A review of research on machine learning for WWTPs}\label{sec:literature}

As noted in Section \ref{sec:introduction}, there is a large variety in the design of WWTPs. This makes it difficult to develop a general machine learning framework for wastewater treatment in general; as does the large variation in the conditions and purpose of the different plants. Moreover, since wastewater treatment is a multi-step process, most studies on machine learning only consider one of a few parts of the full process. In Table \ref{tab:studies_split} a number of studies, including those with the corresponding plants mapped in Figure \ref{fig:map}, are grouped by machine learning method(s) and the application(s) of the modelling. This table and the summary to follow focus on recent studies, chiefly from 2018 and onwards.

\begin{table}[h]
\centering
\caption{The cited studies grouped by the machine learning method(s) used and the object(s) of the modelling. Some of the papers test several methods; in this case, we have grouped them with the best-performing method(s). The methods: \textbf{FNN}: feedforward neural network; \textbf{RNN}: recurrent neural network; \textbf{ELM}: extreme learning machine; \textbf{GB}: gradient boosting; \textbf{AB}: AdaBoost; \textbf{RF}: random forest; \textbf{SVM}: support  vector machine; \textbf{Fuzzy}: fuzzy logic of fuzzy neural network; \textbf{Linear}: linear statistical model; \textbf{Other}: machine learning method not belonging to the other categories.}
\resizebox{\textwidth}{!}{
\begin{tabular}{|c|c|c|c|c|c|}
\hline
 & FNN & RNN & ELM & GB & AB \\ 
\hline
Nitrogen/nitrate/ammonium & \cite{icke2020performance} & \cite{farhi2021prediction, lv2024enhancing, sheng2023exploring, xie2024hybrid, yaqub2020modeling} &  & \cite{yaqub2022modeling} & \cite{zhao2022machine} \\ 
\hline
Phosphate/phosphorous & \cite{wang2021machine} & \cite{yaqub2020modeling} &  & \cite{wang2022towards, yaqub2022modeling} &  \\ 
\hline
BOD/COD & \cite{oliveira2002simulation, nourani2021artificial, el2021forecasting} & \cite{cheng2020forecasting, geng2024multi} & \cite{mekaoussi2023predicting, lotfi2019predicting, liu2020prediction} &  & \cite{sharafati2020potential, zhao2022machine, huang2023construction} \\ 
\hline
TDS/TSS & \cite{wang2021machine, el2021forecasting, gholizadeh2024machine} &  & \cite{lotfi2019predicting} & \cite{wang2022towards} &  \\ 
\hline
Other water quality parameters & \cite{durrenmatt2012data} & & \cite{liu2020prediction} &  & \\ 
\hline
Sludge generation & \cite{newhart2020hybrid, bagheri2015modeling} &  &  &  &  \\ 
\hline
Fault detection &  & \cite{mamandipoor2020monitoring, peng2021effective} &  & \cite{bellamoli2023machine} &  \\ 
\hline
Energy efficiency & \cite{torregrossa2018machine, xu2021integrated, wang2023online, icke2020performance, sadeghassadi2018application}  & \cite{cheng2020forecasting, guo2020data} &  & \cite{bagherzadeh2021prediction} &  \\ 
\hline
% Blank row for spacing (without vertical lines)
\multicolumn{6}{c}{} \\
\hline
 & RF & SVM & Fuzzy & Linear & Other \\ 
\hline
Nitrogen/nitrate/ammonium & \cite{bagherzadeh2021comparative, wu2023coupling} & \cite{guo2015prediction, kim2021machine, zhu2022improved} &  & \cite{vasilaki2018relating} & \cite{yang2022prediction, zaghloul2022application} \\ 
\hline
Phosphate/phosphorous & \cite{wang2021machine} &  &  & \cite{wang2022artificial} & \cite{ly2022exploring} \\ 
\hline
BOD/COD &  & \cite{zhu2022improved} &  &  & \cite{yang2021adaptive, yang2022prediction} \\ 
\hline
TDS/TSS & \cite{wang2021machine} &  &  &  &  \\ 
\hline
Other water quality parameters & \cite{durrenmatt2012data} & \cite{al2024data, granata2017machine} & \cite{han2018data} & \cite{ebrahimi2017temporal, durrenmatt2012data} & \cite{hvala2020design, li2022application} \\ 
\hline
Sludge generation &  &  & \cite{han2018data, han2019data} & \cite{newhart2020hybrid} & \cite{ekinci2023application} \\ 
\hline
Fault detection &  & \cite{harrou2018statistical, dairi2019deep} &  & \cite{xiao2017fault} & \cite{wang2021soft} \\ 
\hline
Energy efficiency & \cite{torregrossa2018machine} &  & \cite{torregrossa2017data, bernardelli2020real, han2018data} &  & \cite{alali2023unlocking} \\ 
\hline
\end{tabular}
}
\label{tab:studies_split}
\end{table}

\subsection{Summary of the studies}

There are several recent review papers that summarises the literature on machine learning of wastewater treatment, while also addressing research needs and highlighting the considerable challenges that need to be addressed before the technology can be fully utilized in daily operations. 
For reviews focusing on explaining relevant concepts within machine learning and the wastewater treatment process, see \cite{alvi2023deep, corominas2018transforming, malviya2021artificial, newhart2019data, singh2023artificial, sundui2021applications}.
For a deeper dive into the literature with comprehensive lists of the ML models tested and the specific target used in each paper, see \cite{bahramian2023data, duarte2023review, jadhav2023water, lowe2022review,  zhao2020application}. The review by Jadhav et al.\ \cite{jadhav2023water} is particularly extensive, although it mostly focuses on neural networks.
Reviews of specific types of models or areas of application include Lu et al.\ \cite{lu2023automatic, matheri2022sustainable} on automatic control, Croll et al.\ \cite{croll2023reinforcement} on reinforcement learning, and Oruganti et al.\ \cite{oruganti2023artificial} on ML for microalgae cultivation and classification.

Among the recent literature we wish to mention, some articles provide a theoretical framework, or a new method, tested only on simulated data, usually obtained with the so-called Benchmark Simulation Model no.\ 1 or no.\ 2 (BSM1 or BSM2) \cite{alex2008benchmark, nopens2010benchmark}; see e.g.\ \cite{han2019data, han2018multiobjective, heo2021hybrid, peng2021effective, sadeghassadi2018application, xiao2017fault, yang2021adaptive}. However, there is also a considerable amount of literature on machine learning applied to data obtained from operational plants; see Figure \ref{fig:map}. Some papers suggest new methods or frameworks also in this case, while many test and compare established ML models. Neural networks are identified as the most studied method \cite{bahramian2023data, corominas2018transforming, malviya2021artificial, oruganti2023artificial, singh2023artificial, zhang2023artificial}, with other popular methods being support vector machines (SVM) \cite{al2024data, bahramian2023data, dairi2019deep, granata2017machine, oruganti2023artificial, zhang2024machine, zhao2022machine, zhu2022improved}, fuzzy logic \cite{bahramian2023data, corominas2018transforming, malviya2021artificial,singh2023artificial}, decision trees \cite{bahramian2023data, oruganti2023artificial}, random forest \cite{oruganti2023artificial, singh2023artificial} and genetic algorithms \cite{bahramian2023data, malviya2021artificial}. Recently, neural network models that take sequential time series data as input, i.e.\ recurrent neural networks (RNN) \cite{bagherzadeh2021prediction, dairi2019deep, guo2020data, peng2021effective, singh2023artificial} and specifically long short-term memory (LSTM) \cite{cheng2020forecasting, el2021forecasting, farhi2021prediction, kovacs2022membrane, lv2024enhancing, mamandipoor2020monitoring, yaqub2020modeling} and gated recurrent unit (GRU) models \cite{cheng2020forecasting, farhi2021prediction}, have emerged as a viable option. Incorporating convolutional layers to catch temporal dependencies, as in a convolutional neural network (CNN), has also been suggested and shown to give good results \cite{farhi2021prediction, guo2020data, li2022application}.

Most recent papers comparing ML models on real-world treatment plant data typically consider some type of a feedforward neural network (FNN), an SVM, and a random forest algorithm.  The XGBoost algorithm is shown to perform best in most of the studies that consider this and compare it to other out-of-the-box ML models on real-world data; see \cite{bellamoli2023machine, sharafati2020potential, wang2022towards, yaqub2022modeling, zhang2024machine}. This is an ensemble model comprising decision trees. Other studies \cite{sharafati2020potential, zhao2022machine} tested the AdaBoost algorithm, a different ensemble decision tree model, and found this to outperform XGBoost (for predicting total dissolved solids (TDS) in the former paper), FNN, SVM and standalone decision tree models (in the latter paper), and random forest (in both papers). On the contrary, Gholizadeh et al.\ \cite{gholizadeh2024machine} found an FNN model to outperform AdaBoost in predicting the total suspended solids (TSS) concentration in the water effluent. Other methods that are deemed to perform best in competition with other ML algorithms, on different targets, are neural network models like FNN \cite{gholizadeh2024machine}, CNN \cite{li2022application} and LSTM \cite{mamandipoor2020monitoring}, linear statistical models like SARIMAX \cite{ly2022exploring}, a kernel ridge regressor \cite{ekinci2023application}, SVMs \cite{al2024data, dairi2019deep, granata2017machine, zhu2022improved} and random forests \cite{ekinci2023application}. This demonstrates the need for testing, as well as developing, machine learning methods tailored for specific processes and data.

Application-wise, the literature most relevant for our paper are those that consider nitrogen removal in some form, although few if any do denitrification in the same way as at Veas. Papers that aim at modelling the removal of nitrate or nitrogen include \cite{bagherzadeh2021comparative, farhi2021prediction, guo2015prediction, kim2021machine, lv2024enhancing, yaqub2020modeling, yaqub2022modeling, yang2022prediction, zaghloul2022application, zhao2022machine, zhu2022improved}. There is also a lot of literature on modelling other specific aspects of the process that may pose similar methodological challenges for the machine learning, like the concentration of other chemicals \cite{bagheri2015modelingb, granata2017machine, han2018data, huang2023construction, liu2020prediction, ly2022exploring, mekaoussi2023predicting, newhart2020hybrid, newhart2020hybrid, nourani2021artificial, sun2017reduction, wang2022artificial, yang2021adaptive}, the concentrations of solids \cite{bagheri2015modelingb, gholizadeh2024machine, lotfi2019predicting, sharafati2020potential, wang2022towards, wang2021machine}, the total energy consumption \cite{alali2023unlocking, bagherzadeh2021prediction, torregrossa2017data, torregrossa2018machine} or sludge generation \cite{bagheri2015modeling, ekinci2023application, zhang2024machine}. There are also some studies that consider a specific plant and aim to generate a model for many parameters or the complete process \cite{al2024data, cheng2020forecasting, el2021forecasting, icke2020performance}.

A reliable predictive ML model can be used for decision support, but the ultimate use of a data-driven model is in closed-loop control of the operation. That means considering how different ML techniques integrates with different strategies and techniques within control theory. Papers that take this perspective include \cite{bernardelli2020real, cao2020online, guo2020data, han2018data, han2018multiobjective, heo2021hybrid, sadeghassadi2018application, zhang2022improved}.

\subsection{Challenges and future directions}

Much of the literature, both the review papers and the case studies, point to many of the same challenges with data-driven learning of wastewater treatment processes. The large gap between research and industrial applications is highlighted in many studies. A close collaboration between academia, research institutes, the private sector and the public sector is seen as necessary for future developments \cite{bahramian2023data, lowe2022review}. Some also point to the lack of association between engineering sciences and the machine learning community as a persisting issue \cite{corominas2018transforming, newhart2019data}.

On a practical level, the lack of high-quality training data is the big, overarching issue \cite{alvi2023deep, corominas2018transforming, jadhav2023water, lowe2022review, matheri2022sustainable, oruganti2023artificial}. The ever-changing states of the process due to weather conditions and the considerations of the operations make it difficult to generalize data-driven models \cite{alvi2023deep, jadhav2023water, lowe2022review}, as does the lack of consensus on the design procedure for optimal control and operation of WWTPs \cite{lu2023automatic, newhart2019data}. 

Moreover, since the models are to be used by operators at the plants, explainability and interpretability is of high importance \cite{alvi2023deep, duarte2023review, lowe2022review}. Trustworthy AI is a field in itself, and water treatment is certainly an area where this will be given high priority in the coming years. The end-users both inside and outside of the treatment facilities need to trust the machine learning models used, not just the developers of the algorithms. The models need to be robust against fluctuations in the data, both due to sensor variations and changing conditions. Unpredictable and unstable behaviour outside the domain of the training data is a common challenge with machine learning models and needs to be addressed for the technology to be seen as reliable by those who will use it \cite{alvi2023deep}.

To solve some of these challenges, many papers point to hybrid models that combine mechanistic and data-driven models \cite{alvi2023deep, duarte2023review, jadhav2023water, newhart2019data, singh2023artificial}. Schneider et al.\ \cite{schneider2022hybrid} review the status and opportunities with this approach in both wastewater and drinking water treatment. With some notable exceptions \cite{hvala2020design, wang2023online, wu2023coupling}, few concrete hybrid models have been proposed and tested out on data from treatment plants. Given the considerable variations between plants and processes, there is much to be gained from putting more effort into developing hybrid models. In that regard, the software used for modelling today need to be amended to work well in combination with data-driven learning \cite{croll2023reinforcement, jadhav2023water}.

\renewcommand{\thefigure}{C.\arabic{figure}}
\setcounter{figure}{0}

\section{Additional figures}\label{sec:app:plots}

\begin{figure}
    \centering

    \begin{subfigure}[b]{\textwidth}
        \centering
        \includegraphics{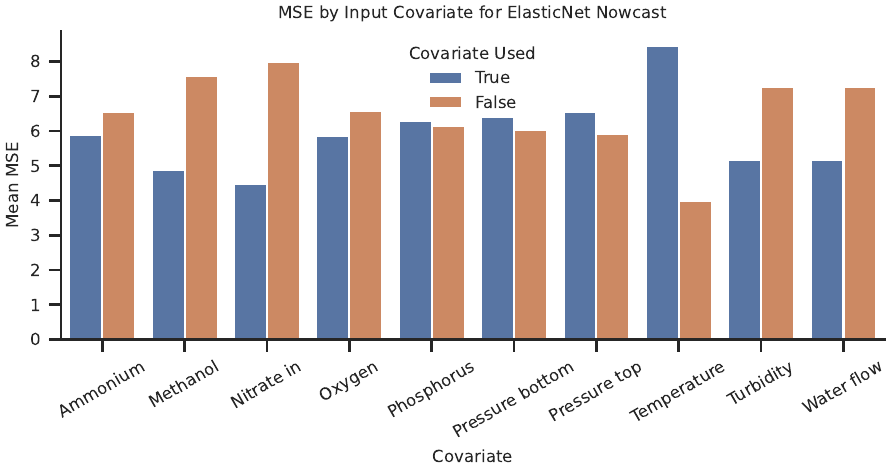}
        \label{fig:inputs:nowcast:elastic_net:onoff}
    \end{subfigure}
    \begin{subfigure}[b]{\textwidth}
        \centering
        \includegraphics{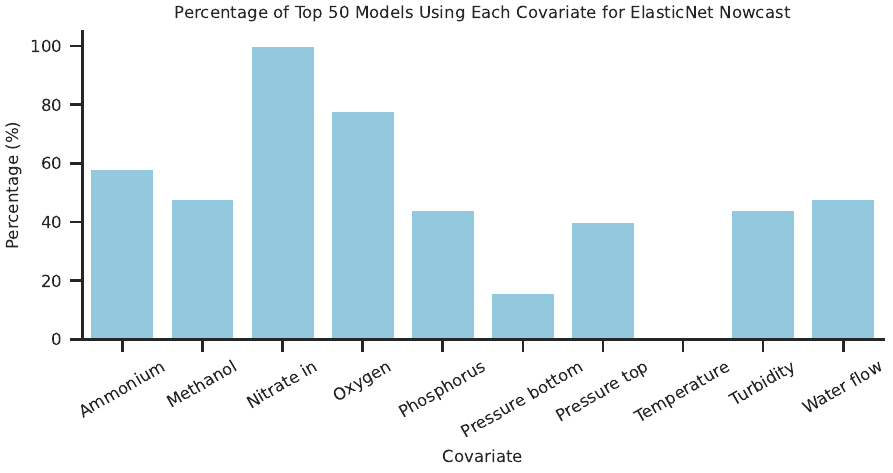}
        \label{fig:inputs:nowcast:elastic_net:top}
    \end{subfigure}

    \caption{Covariate contributions for the Elastic Net model.}
    \label{fig:inputs:elastic_net}
\end{figure}

\begin{figure}
    \centering

    \begin{subfigure}[b]{\textwidth}
        \centering
        \includegraphics{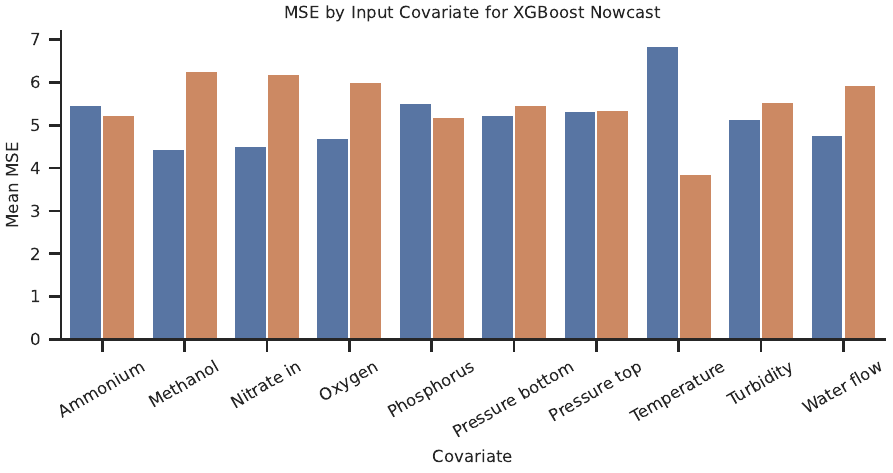}
        \label{fig:inputs:nowcast:xgboost:onoff}
    \end{subfigure}
    \begin{subfigure}[b]{\textwidth}
        \centering
        \includegraphics{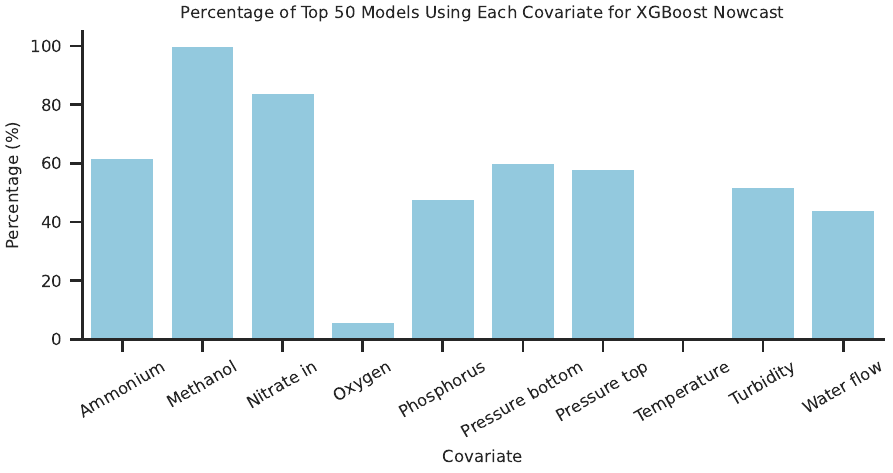}
        \label{fig:inputs:nowcast:xgboost:top}
    \end{subfigure}

    \caption{Covariate contributions for the XGBoost model.}
    \label{fig:inputs:xgboost}
\end{figure}
    
\begin{figure}
    \centering

    \begin{subfigure}[b]{\textwidth}
        \centering
        \includegraphics{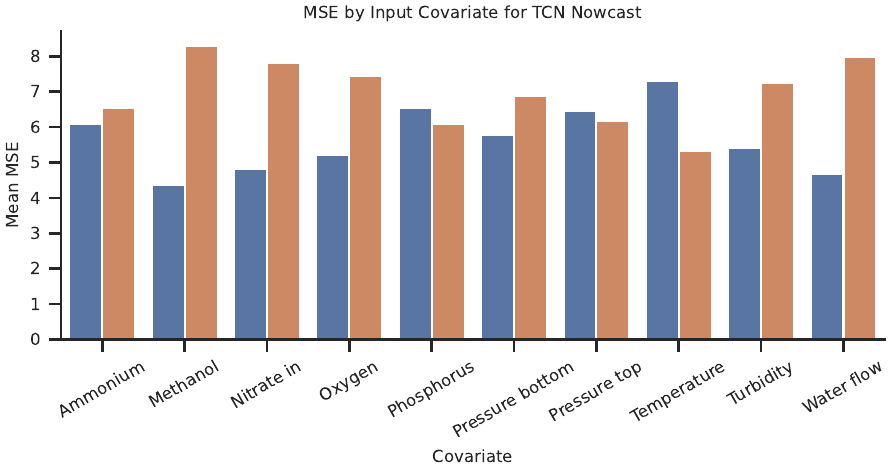}
        \label{fig:inputs:nowcast:tcn:onoff}
    \end{subfigure}
    \begin{subfigure}[b]{\textwidth}
        \centering
        \includegraphics{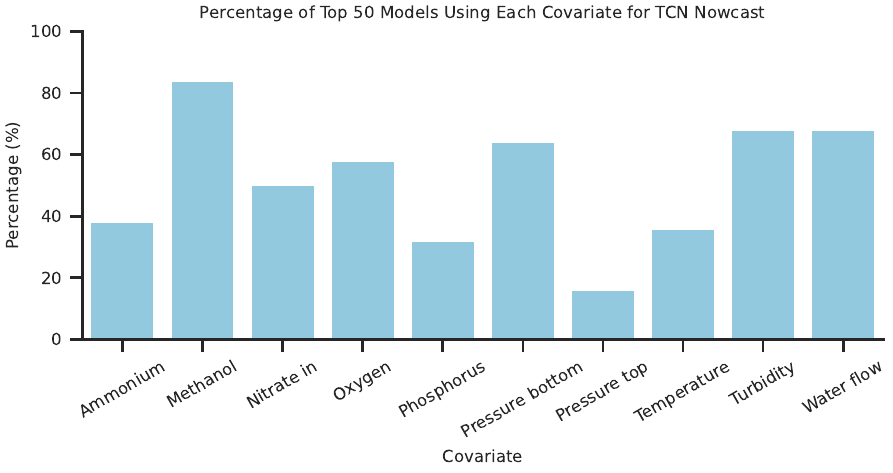}
        \label{fig:inputs:nowcast:tcn:top}
    \end{subfigure}

    \caption{Covariate contributions for the TCN model.}
    \label{fig:inputs:tcn}
\end{figure}

\begin{figure}[htbp]
    \centering
    \includegraphics{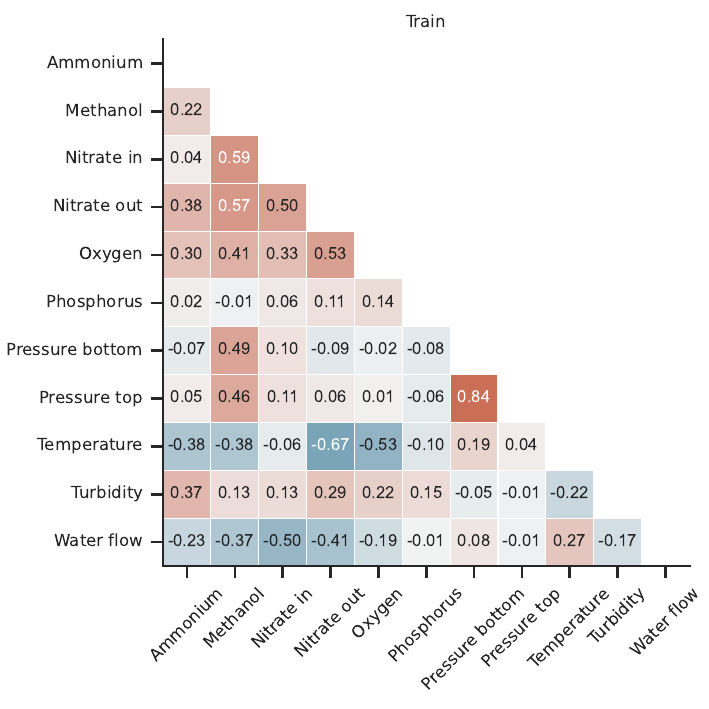}
    \caption{Feature correlations for all features on the train dataset.}
    \label{fig:feature_correlations:train}
\end{figure}

\begin{figure}[htbp]
    \centering
    \includegraphics{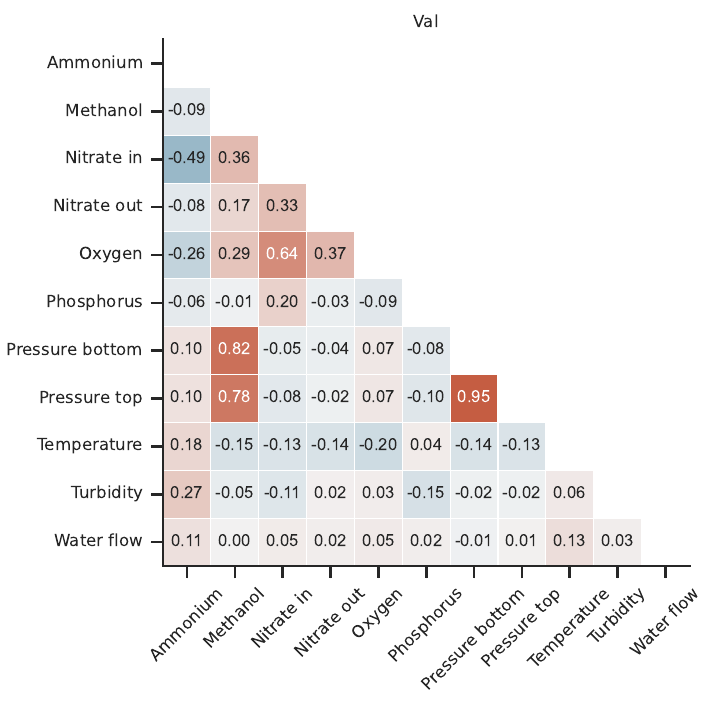}
    \caption{Feature correlations for all features on the val dataset.}
    \label{fig:feature_correlations:val}
\end{figure}

\begin{figure}[htbp]
    \centering
    \includegraphics{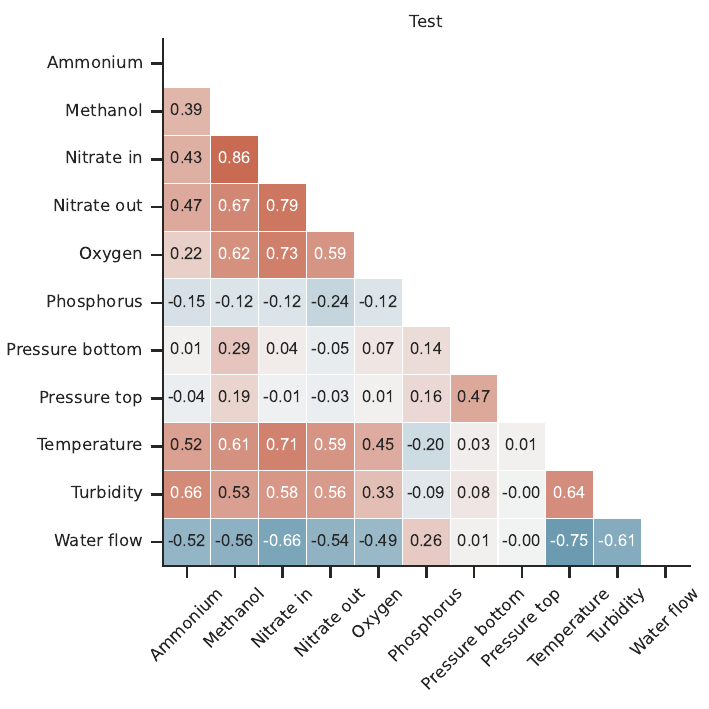}
    \caption{Feature correlations for all features on the test dataset.}
    \label{fig:feature_correlations:test}
\end{figure}
\end{document}